\pdfoutput=1

\documentclass[11pt]{article}

\usepackage[final]{acl}

\usepackage{times}
\usepackage{latexsym}

\usepackage[T1]{fontenc}

\usepackage[utf8]{inputenc}

\usepackage{microtype}

\usepackage{inconsolata}
\PassOptionsToPackage{dvipsnames,svgnames,table}{xcolor}
\usepackage[dvipsnames]{xcolor}
\usepackage{multirow}
\usepackage{pifont}
\usepackage{longtable}
\usepackage{tabularx}
\usepackage{booktabs}
\usepackage{pifont}
\usepackage{xspace}
\usepackage{subfig}
\usepackage{subcaption}
\usepackage{lipsum}
\usepackage{hyperref}
\usepackage{url}
\usepackage{pdfpages}
\usepackage{soul}
\usepackage{booktabs}
\usepackage{makecell}
\usepackage{twemojis}

\newcommand*{\affaddr}[1]{#1} 
\newcommand*{\affmark}[1][*]{\textsuperscript{#1}}

\newcommand{\add}[1]{\textcolor{black}{#1}}

\makeatletter
\def\thanks#1{\protected@xdef\@thanks{\@thanks
        \protect\footnotetext{#1}}}
\makeatother

\usepackage{graphicx}
\usepackage{array}
\usepackage[table,xcdraw]{xcolor}
\usepackage{booktabs}
\usepackage{adjustbox}
\usepackage{enumitem}
\usepackage{float}
\newfloat{promptboxfloat}{tbp}{lop}
\floatname{promptboxfloat}{Prompt}

\usepackage[most]{tcolorbox}
\usepackage{multicol}
\usepackage{fontawesome5}
\tcbset{
  promptbox/.style={
    enhanced,
    colback=blue!5,
    colframe=blue!40!black,
    boxrule=0.5pt,
    arc=2pt
  }
}

\definecolor{FCE8E8}{HTML}{FCE8E8} 
\definecolor{E7F3E6}{HTML}{E7F3E6} 

\usepackage{highlight}
\usepackage{cleveref}

\newcommand\numannotated{284\xspace}
\newcommand\numsampled{356\xspace}
\newcommand\totalquestions{37\xspace}
\newcommand\corequestions{20\xspace}
\newcommand\otherquestions{17\xspace}

\newcolumntype{L}[1]{>{\raggedright\let\newline\\\arraybackslash\hspace{0pt}}p{#1}}
\newcolumntype{R}[1]{>{\raggedleft\let\newline\\\arraybackslash\hspace{0pt}}p{#1}}
\newcolumntype{M}[1]{>{\raggedright\let\newline\\\arraybackslash\hspace{0pt}}m{#1}}

\definecolor{darkgreen}{rgb}{0.0, 0.4, 0.13}
\definecolor{darkgreen2}{HTML}{037171}
\definecolor{lightgreen2}{HTML}{83c5be}
\definecolor{coral2}{HTML}{e29578}
\definecolor{lightlightblue}{rgb}{0.9, 0.95, 1.0}

\definecolor{mustard}{rgb}{0.9, .61, .11}

\newcommand{\greenhl}[1]{%
  \colorbox{darkgreen2}{\textcolor{white}{#1}}%
}

\newcommand{\lightgreenhl}[1]{%
  \colorbox{lightgreen2}{\textcolor{black}{#1}}%
}
\newcommand{\coralhl}[1]{%
  \colorbox{coral2}{\textcolor{black}{#1}}%
}

\title{Illusions of the Gold Standard: A Large-scale Analysis of \\Human Evaluation Protocols for Long-form Text Generation}

\author{%
Katelyn Xiaoying Mei\affmark[1]\Thanks{ Denotes equal contribution} \quad
Yi-Li Hsu\affmark[1,2]\footnotemark[1] \quad 
Minjoon Choi\affmark[3]\quad
Zongwan Cao\affmark[1]\\ 
\textbf{Chenjun Xu\affmark[1]\quad
Bingbing Wen\affmark[1]\quad
Su Lin Blodgett\affmark[4]\quad
Lucy Lu Wang\affmark[1,5]}\\ [2mm]
\affaddr{\affmark[1]University of Washington}\quad
\affaddr{\affmark[2]National Tsing Hua University} \\
\affaddr{\affmark[3]Seoul National University} \quad
\affaddr{\affmark[4]Mila - Qu\'{e}bec AI Institute} \quad
\affaddr{\affmark[5]Allen Institute for AI}\\ [0.5mm]
\texttt{\{kmei, lucylw\}@uw.edu}
}

\begin{document}
\maketitle
\begin{abstract}
Human evaluation plays a critical role in assessing the quality of generated text. However, the reliability and reproducibility of these evaluations depend on transparent and well-documented protocols---details that are frequently missing in current practice. In this work, we conduct a large-scale analysis of human evaluation protocols for evaluating long-form generation tasks in *CL conference publications from 2023--2025, including a full manual review of \numannotated papers and LLM-assisted analysis for another 1.8k+ papers. We define a set of \corequestions reportable criteria related to reproducibility of human evaluation studies, and apply these criteria to systematically examine reporting norms and practices within the community. We find widespread under-reporting of important aspects of human evaluation study design, leading to ambiguity about what was measured and how, who contributed judgments, and how judgments should be interpreted. Based on these findings, we outline actionable recommendations to support more transparent and reproducible reporting in future research.\footnote{Our analysis code and annotated dataset can be found at: \href{https://github.com/larchlab/Illusions-of-the-Gold-Standard}{https://github.com/larchlab/Illusions-of-the-Gold-Standard}.}

\end{abstract}

\section{Introduction}
\label{sec:intro}

With the growing adoption of LLMs, long-form or open-ended generation now dominates NLP research.\footnote{In our analysis, around half of *CL papers from 2025 study long-form generation tasks per our definition (Table~\ref{tab:longform_proportion}).} Automated metrics work poorly in these settings and human evaluation of model performance is still considered the gold standard, especially in high-expertise domains such as healthcare~\cite{fraile2025expert, wang-etal-2023-automated}, science~\cite{idahl-ahmadi-2025-openreviewer}, law~\cite{fei2025internlm}, policy~\cite{rivera2024escalation}, and misinformation detection and mitigation~\cite{mishra2024fine,cho2024can}. Yet, as prior work has shown for specific domains and/or tasks \cite{awasthi2025human}, current human evaluation procedures lack proper standardization and operationalization, which can limit the validity of evaluation, the robustness of conclusions, comparability across studies, and reproducibility of evaluation findings \cite{elangovan2024considers,fleisig-etal-2024-perspectivist}. 

Concurrently, as the (self-)evaluative capabilities of models improve \cite{madaan2023self}, human evaluation is increasingly supplemented and/or supplanted by LLM-judges \cite{bavaresco-etal-2025-llms,thakur2025judging,posner2025judge}. In these cases, human evaluations play an additional and vital meta-evaluative role in assessing the performance of LLM-judges. As the landscape evolves, human judgments remain a foundational part of our evaluation methodology, and should be held to similar rigor and standards as other research methods employed by our community.

With this motivation, we turn the research lens upon the *CL research community to understand reporting practices around human evaluation and to identify and critique shortcomings of current practice. Drawing from prior work on scientific reproducibility and good study design and reporting  \cite{munafo2017manifesto,fleisig-etal-2024-perspectivist}, we define 20 reportable criteria for human evaluation protocols, covering aspects of task definition, annotation operationalization, annotator information, and data analysis and interpretation. We focus on papers studying medium- and long-form natural language generation, as they have high burden for human evaluation, and lack clear and reproducible automated evaluation metrics~\cite{xu-etal-2023-critical}. Through manual and LLM-assisted analysis of this generation literature, we examine reporting patterns for human evaluations as they have evolved over the last few years (2023--2025) according to our criteria.

Our findings highlight clear deficiencies in documentation, with systematic under-reporting of key aspects of human evaluation protocols. For example, only around half of papers we analyze include guidelines for human evaluation tasks or provide justification for the dimensions that were evaluated. And perhaps unsurprisingly, good practices such as the use of power analysis for computing sample size or reporting statistical confidence are exceedingly rare. We also find that the proportion of papers using human evaluation for long-form generation evaluation is declining in recent years, while the proportion of papers using LLM-judges for evaluation appears to be increasing. Based on these results, we offer recommendations for how to improve reporting.

In sum, we contribute the following:
\begin{itemize}[noitemsep, topsep=0pt, leftmargin=10pt]
    \item We define a set of \corequestions core reportable criteria for human evaluation studies along the dimensions of task documentation, annotation design, and analysis and interpretation. In \S\ref{sec:codebook}, we describe the formation of these criteria, our codebook for assessing them, and additional reportable elements associated with good study design;
    \item Using our criteria, we conduct a large-scale analysis of 9.1k+ papers published at *CL conferences from 2023--2025. Of over 1,800 papers that study long-form generation and include human evaluation, we sample 356 papers and manually annotate \numannotated papers in full and conduct LLM-assisted annotation of the remainder. \S\ref{sec:methods} describes our methods for corpus construction and data sampling, and implementation of our human annotation protocol; 
    \item Our analysis (\S\ref{sec:results}) reveals pervasive under-reporting of important aspects of human evaluation, entrenched but poorly justified norms around evaluation design, and recent changes in evaluation and meta-evaluation practices. We provide our recommendations in \S\ref{sec:discussion}.
\end{itemize}

\begin{figure*}
    \centering    \includegraphics[width=1\linewidth]{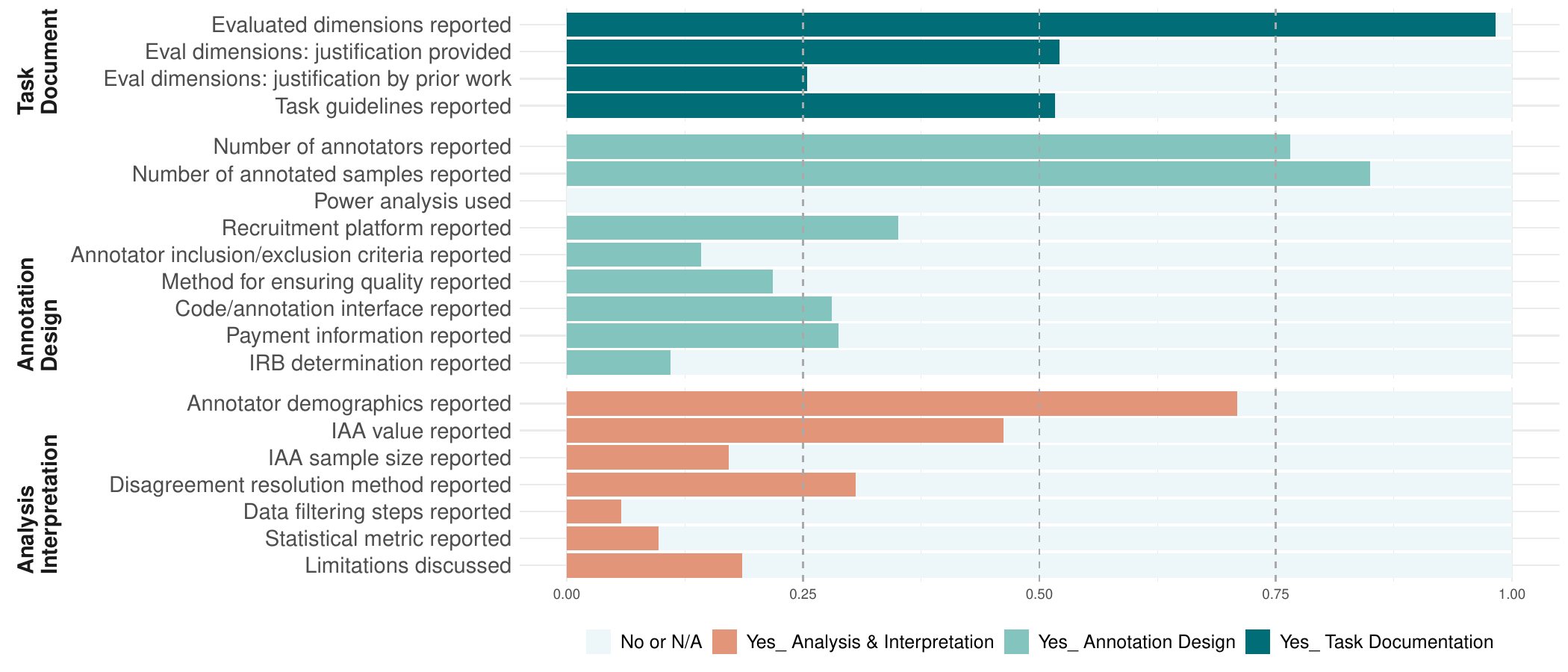}
    \caption{Average proportion of *CL papers reporting each of \corequestions core criteria related to the reproducibility of human evaluation protocols, estimated via bootstrapping. While most papers report evaluation dimensions, some annotator information, and annotation sample size, there is significant under-reporting of other aspects of evaluation study design. The bootstrapped standard deviations for all criteria fall in the range 0.01-0.03.}
    \label{fig:main}
\end{figure*}

\section{Related Work}
\vspace{-1mm}
\label{sec:rw}

\paragraph{Reproducibility in ML/NLP}
Across scientific fields, poor study design and reporting have contributed to a broader reproducibility crisis. These concerns extend to machine learning and NLP, where methodological complexity and differences across studies can make it difficult to replicate results~\cite{howcroft2020twenty,belz2023missing,thomson2024common}.  The community has introduced formal reproducibility frameworks (e.g., conference checklists \cite{dodge-etal-2019-show}, evaluation sheets~\cite{shimorina2022human,belz2025heds}, resource tracks, and structured documentation such as model and data cards \cite{mitchell2019model,rogers-etal-2021-just-think}) to standardize author disclosures about data, experimental design, and evaluation pipelines~\cite{elangovan2024considers}. Yet despite adoption, these frameworks have not fully addressed underlying problems. Checklists are completed by authors without external validation, and their accuracy or completeness is rarely audited during peer review; as a result, important aspects of study design and evaluation can and do go unreported. Our study complements prior work by not only contributing a framework for assessing human evaluation study reporting, but also conducting a large-scale analysis of current reporting practices and their implications for reproducibility.

\vspace{-1mm}
\paragraph{Human Evaluation for Generation Tasks}
Human evaluation has generally been considered the ``gold standard'' evaluation for natural language generation~\cite{celikyilmaz2020evaluation}. Human evaluation methods usually focus on intrinsic evaluation that assesses the quality of LLM-generated text, via data collection approaches like pairwise comparison (i.e., annotators indicate the response they prefer, perhaps along a specific dimension) or scoring scales. 

In recent years, with emergent LLM-as-judge capabilities, human evaluation is also increasingly employed in a meta-evaluative capacity \cite{madaan2023self, bavaresco-etal-2025-llms,posner2025judge,thakur2025judging}.
Given the importance of human evaluation, and critiques \citep{howcroft2020twenty} and anecdotes of under-reporting,
we focus our analysis on recent *CL literature and aim to understand current practices for use of human evaluation in generation tasks.

\section{Reporting Criteria \& Codebook}
\label{sec:annotation}
\vspace{-1mm}
\label{sec:codebook}
We define reportable criteria for human evaluation grounded in the reproducibility literature. 
These criteria capture author-reportable aspects of human evaluation design, data collection, and analysis---the core operational stages of the scientific method \cite{munafo2017manifesto}. We develop these criteria and the associated codebook by mapping each stage to concrete reporting decisions made by authors, and further revise our codebook using insights from extant investigations of human evaluation pitfalls in NLP research. \add{For example, \citet{fleisig-etal-2024-perspectivist} advocate for more detailed reporting of  disagreements when they occur in annotation; in response, we go beyond documenting annotator agreement metrics and also document whether authors provide information about if and how disagreements are resolved.}

\vspace{-1mm}
\paragraph{Codebook development} Two lead authors and two senior authors developed the criteria and codebook iteratively via a combination of inductive and deductive processes~\cite{fereday2006demonstrating}.  
Over four iterations, two authors independently applied initial versions of the codebook to human evaluation protocols from a sample of five papers per iteration. Following each round of coding, the authors discussed and resolved disagreements and refined the codebook by either revising existing codes or introducing new inductive codes. At the end of this process, the study team reached consensus on the suitability of the final codebook. 

The final codebook contains \totalquestions questions, \corequestions of which form our core set of reportable criteria. The other \otherquestions questions collect additional detailed information from each paper.
The final set of codes and answer options is included in Appendix~\ref{app:complete_criteria_list}. 

\vspace{-1mm}
\paragraph{Reportable criteria} The \corequestions core criteria are binary, e.g., a paper either reports the number of annotators (Yes), or no information is provided (No or N/A). We group them into the following categories for presentation (see Figure~\ref{fig:main} for full list):

\begin{itemize}[leftmargin=10pt, topsep=1pt, itemsep=-2pt]
    \item \greenhl{Task documentation (4 criteria)}:~Aspects related to what is evaluated and how it is described to annotators, e.g., what dimensions (preference, accuracy, factuality, etc.) are being evaluated, justification for these dimensions, and guidelines for how annotators should judge each dimension.
    \item \lightgreenhl{Annotation design (9 criteria)}:~Operational details of the annotation procedure, such as the annotation interface, sample size, and processes ensuring annotation quality; as well as the annotators, e.g., recruitment platform, inclusion/exclusion criteria, payment, number of annotators.
    \item \coralhl{Analysis \& interpretation (7 criteria)}:~Aspects related to how collected annotations are analyzed, interpreted, and presented. Results from human evaluation can vary due to (i) variance in annotator judgment and (ii) variance in methods used to analyze the collected data. 
    This category therefore includes elements such as annotator demographics, agreement (i.e., interrater reliability),
    whether additional data processing steps are used prior to reporting results (disagreement resolution or data filtering), reporting of statistical metrics, and whether limitations are discussed.
\end{itemize}

\noindent Not all \corequestions criteria apply in every setting, e.g., if disagreements are resolved by discussion, then reporting IAA may be unnecessary. Nevertheless, we expect most studies involving human evaluation to report roughly 15-16 criteria from this list.

\vspace{-1mm}
\paragraph{Other information}

Beyond the core reportable criteria, we also collect \otherquestions additional pieces of information from each paper for further analysis. First, we collect information that helps obtain a more nuanced understanding of evaluation design and reporting practices, such as where human evaluation details are reported (in the main paper or appendix), and whether LLMs and humans are used to evaluate the same dimensions (to understand the increasing use of LLM-judges in evaluation). Second, we collect details reported about annotators to better characterize annotator populations, including whether they are students, experts, or authors, and the platforms from which they are recruited. Last, we track details of IAA metrics such as the specific metrics computed and any methods for resolving disagreements (e.g., majority vote or consensus processes). Refer to Appendix~\ref{app:complete_criteria_list} for complete codebook details.

\section{Dataset \& Methods}
\vspace{-0.5mm}
\label{sec:methods}

\begin{table*}[ht]
\footnotesize
\centering
\renewcommand{\arraystretch}{0.82}

\begin{adjustbox}{max width=\textwidth}
\begin{tabular}{l l r r r r r}
\toprule
\textbf{Year} & \textbf{Conference} &
\multicolumn{5}{c}{\textbf{Counts (proportion)}} \\
\cmidrule(lr){3-7}
& & \textbf{Total} &
\makecell{\textbf{Step 1}\\\textbf{(Keywords):}\\\textbf{Longform}} &
\makecell{\textbf{Step 2A}\\\textbf{(LLM Filter):}\\\textbf{Longform}}  &
\makecell{\textbf{Step 2B (LLM}\\\textbf{ Filter): Longform}\\\textbf{\& Human Eval}} &
\makecell{\textbf{Step 3: Sample}\\\textbf{for Manual }\\\textbf{Annotation}} \\
\midrule
2023 & EACL  &  281 &  225 (0.80) &  65 (0.23)  &  39 (0.14) & - \\
     & ACL   &  912 &  865 (0.95) & 289 (0.32)  & 192 (0.21) &  - \\
     & AACL  &   73 &   64 (0.88) &  22 (0.30)  &  19 (0.26) & - \\
     & EMNLP & 1048 &  919 (0.88) & 355 (0.34)  & 196 (0.19) & - \\
\midrule
2024 & NAACL &  489 &  452 (0.92) & 153 (0.31)  & 105 (0.21) & 20 \\
     & ACL   &  869 &  793 (0.91) & 275 (0.32)  & 177 (0.20) & 135 \\
     & EMNLP & 1270 & 1156 (0.91) & 381 (0.30) & {256} (0.20) & 81 \\
     & EACL  &  182 &  161 (0.89) &  62 (0.34)  &  35 (0.19) & 20 \\
\midrule
2025 & NAACL &  637 &  589 (0.92) & 300 (0.47) & 145 (0.23) & 30 \\
     & ACL   & 1602 & 1517 (0.95) & 806 (0.50) & 351 (0.22)& 70 \\
     & EMNLP & 1809 & 1667 (0.92)& 912 (0.50) & 376 (0.21)& - \\
\midrule
 & {Total} & {9172} & {8408 (0.92)} & {3620 (0.39)} &{1891 (0.21)} &{356} \\

\bottomrule
\end{tabular}%
\end{adjustbox}
\caption{Progressive filtering of *CL papers for inclusion in analysis. Keyword filters (step 1) are followed by LLM filters for papers about long-form generation tasks (step 2A) \textbf{and} which contain human evaluation (step 2B). We then stratify sample over 6 conferences from 2024--2025 for manual annotation (step 3). Proportions represent per-row normalization. Conferences are ordered chronologically according to their official event dates.}

\label{tab:longform_proportion}
\vspace{-1mm}
\end{table*}

Using our criteria and codebook, we conduct a large-scale manual and LLM-assisted analysis of *CL conference publications from 2023--2025. We focus on these venues because they constitute a coherent and influential publication ecosystem for computational linguistics and NLP while spanning a diverse range of research, offering a representative snapshot of prevailing evaluation practices. We analyze papers from the last three years to capture current practices during a period of rapid change: (i) the growth of LLMs enables new generation tasks; (ii) unlike tasks that can be assessed with automated metrics, medium- and long-form generation tasks lack reference answers and have higher evaluative burden; and (iii) the prevailing use of human evaluation for direct assessment and LLM-judge meta-evaluation raises questions about how human evaluation protocols should be designed.

Papers are included in our analysis if they meet our inclusion criteria: (i) studies a long-form generation task and (ii) employs human evaluation. We define \textit{long-form generation} as free-form natural language generation, excluding tasks such as machine translation and code generation. We define \textit{human evaluation} as the use of human annotators to examine and assess model outputs. 
A subset of \numannotated papers is manually annotated in full using our codebook from \S\ref{sec:codebook}, while we conduct partial analysis on the remaining papers through LLM-assisted labeling. 
Below, we describe our procedures for corpus curation and sampling (\S\ref{sec:dataset}), manual annotation (\S\ref{sec:annotation}), and LLM-assisted labeling (\S\ref{sec:simulation}). 

\subsection{Corpus Curation}
\label{sec:dataset}

We begin with 9,172 papers from major *CL conferences: ACL, EMNLP, and regional chapters NAACL, EACL, and AACL, published 2023--2025. We download conference proceeding paper PDFs from the ACL Anthology\footnote{\href{https://aclanthology.org/}{https://aclanthology.org/}} and use GROBID\footnote{\href{https://grobid.readthedocs.io/en/latest/}{https://grobid.readthedocs.io/en/latest/}} following \citet{acl_anthology_corpus} to parse and extract clean textual content for downstream filtering and search.

\vspace{-0.5mm}
\paragraph{Step 1: Keyword filters}
We first narrow the corpus to papers studying long-form text generation. We expand a set of seed keywords (summarize/summarise, dialogue, long-form, etc.) with GPT-4 to cover related task variants (e.g., multi-turn dialogue, document-level generation). We apply case-insensitive matching with stemming across titles, abstracts, and the main text of each paper, retaining papers that match at least one expanded keyword. This produces a candidate set of 8,408 papers. We provide the full keyword filter set in Appendix~\ref{app:keyword_filters}.

\vspace{-0.5mm}
\paragraph{Step 2: LLM filters}
We apply a second-stage filter via majority vote among three LLMs: Gemini-2.5-Pro, Claude-3.7-Sonnet-20250219, and GPT-4o-mini-2025-04-16. Each model answers two binary screening questions: (i) whether the task in the paper is considered long-form natural language generation and (ii) whether the paper involves human evaluation. A paper is retained if at least two models answer ``Yes'' for both criteria. 
Following this step, 3,620 papers meet our inclusion criteria for long-form generation, and 1,891 papers further meet our criteria of using human evaluation. Prompts for filtering can be found in Appendix~\ref{app:llm_filters}.

To estimate the false negative rate of these LLM filters, we manually inspect a random sample of 50 papers that the LLM majority vote finds to be about long-form generation but rejects for not including human evaluation. Among these 50 papers, 3 are found to include human evaluation, corresponding to a FNR of around 6\%.

\vspace{-1mm}
\paragraph{Step 3: Sampling for manual annotation} We sample 356 papers from the set of 1,891 for manual annotation. We sample only from conferences in 2024 and 2025 to focus our efforts on capturing recent practices.
As such, we stratify our sample to include approx.~20\% of papers from all 4 conferences in 2024 and NAACL and ACL in 2025.\footnote{EMNLP 2025 occurred after we completed manual annotations, though it is included in automated analysis.}$^,$\footnote{Our final manual annotation set includes proportionally more papers from some conferences. We had planned to annotate more papers (before recalibrating due to the difficulty of the task) and our initial annotation assignments did not randomize conference order. We corrected for this midway through our annotation timeline to achieve more balanced coverage over all included conferences.} 

\vspace{-1mm}
\subsection{Manual Annotation Procedure}
\label{sec:annotation}
\vspace{-1mm}
\label{sec:annotation_methods}

Two lead authors and three contributors coded the \numannotated sampled papers. The lead authors are experienced NLP and HCI researchers familiar with reading research papers, and the three contributors---one undergraduate and two masters students---have prior experience reading and writing NLP papers. 

\vspace{-1mm}
\paragraph{Annotation process}
To operationalize our annotation task, we provided each annotator with a codebook reference sheet with definitions for each code and how to select different answer options (Appendix~\ref{app:codebook_reference}). \add{To reduce the need for subjective interpretation, our annotation instructions are deliberately minimal and lenient, intended to capture overall reporting trends; we give credit when a paper reports \emph{any} information about an item and do not judge whether what is reported is sufficient.} 

The annotation task was conducted using Google Sheets (see Appendix~\ref{app:annotation_interface} for annotation interface). Each batch of papers was assigned to an annotator in a new tab in their own spreadsheet. Within the interface, we grouped the \totalquestions total questions answered for each paper together into subcategories likely to be documented in the same paper sections. We restricted all binary and multiple-choice questions to a predefined set of answer options (pre-configured in the spreadsheet); where appropriate, annotators may also select ``Other'' and provide a free-text explanation.
In cases where multiple human evaluation protocols are described in the same paper, annotators were instructed to focus on the first protocol described in the paper. 

\vspace{-1mm}
\paragraph{Onboarding} To ensure all contributors had a comprehensive understanding of the codebook and task, we began with a three-week training period consisting of: (i) an introductory group meeting to explain the codebook; (ii) an initial batch of 10 papers annotated independently by all contributors; (iii) comparing annotation results to the leads and one-on-one discussions to provide feedback and clarify disagreements (during these discussions, we also refined definitions for any ambiguous codes); and (iv) a second independent round of annotation of 5 papers to reassess performance. 

All collaborators reached 73\% agreement after iterations of feedback and were assigned their own non-overlapping batches of papers each week for annotation. Each annotator annotated between 50 and 138 papers over the study period.
Furthermore, one of the lead authors conducted a random quality check on 105 (60\%) papers of other annotators.

\vspace{-1mm}
\paragraph{Inter-annotator~agreement (IAA)}
We compute IAA for all annotators using 5 papers from the final onboarding annotation set. There are 155 questions per person (115 binary, 30 single-label multiple choice (MC), and 10 multi-label MC; open-ended questions not included in IAA computation).
We report average pairwise agreement between each of the three contributing annotators and the consensus annotation provided by the two lead authors.

Binary questions (n=23) yield the most reliable judgments (percent agreement=81\%; Cohen’s $\kappa$=0.51), while single-label MC questions (n=6) achieve fair agreement (percent agreement=69\%; Cohen’s $\kappa$=0.52). Multi-label MC questions (n=2) also achieve fair agreement (percent agreement=53\%; Cohen’s $\kappa$=0.46). 

\vspace{-1mm}
\paragraph{Data analysis \& interpretation}

We construct additional binary dummy variables from questions that have numeric, MC, or qualitative answers. For example, from the numeric IAA value, we create a new binary variable \texttt{IAA\_reported} that is coded ``Yes'' if a specific number is reported in the paper, and ``No or N/A'' otherwise. To estimate the true proportion of papers that report each criterion, we perform bootstrap resampling with replacement (n=500) to derive averages and standard errors.

\vspace{-1mm}
\subsection{LLM-assisted Annotation}
\label{sec:simulation}
\vspace{-0.5mm}

To scale analysis to our whole corpus, we adopt LLMs to label papers we could not manually annotate. For each paper, we construct its input context from its abstract, introduction, and candidate human evaluation sections from its main paper and appendix identified using keywords (Appendix~\ref{app:human-eval-section-keywords}).
We then prompt GPT-4o-mini-2025-04-16 with codebook questions to extract relevant information about the first human evaluation pipeline reported in each paper. 
The model is prompted with a chunk of questions from our codebook at a time.
We conduct automatic type checking and apply numeric-or-NA constraints; if chunk-level or whole-paper validation fails, we re-run the full set of extraction prompts up to two more times. 

For prompt refinement, we use a training set of 26 papers sampled from those manually 
annotated by the two lead authors. We test final prompt performance on an independent validation set of 125 papers evenly sampled from all five independent annotators (25 papers each). Model selection and prompting details can be found in Appendix~\ref{app:llm-annotation-details}.

In \S\ref{sec:results}, we only report
LLM annotation results for questions where the LLM 
achieves a validation accuracy over 0.75.
Validation accuracies for all binary and multiple-choice questions and final prompts are reported in Appendix~\ref{app:llm_annotation_prompts}.

\section{Results}
\label{sec:results}
\label{sec:results_human}
\vspace{-1mm}
\paragraph{Summary statistics} 
Among \numsampled manually-annotated papers, \numannotated are confirmed by annotators to meet both inclusion criteria: (i) studying a long-form generation task and (ii) including a human evaluation pipeline; i.e., 72 papers did not meet either one or both of these criteria, and can be considered false positives from the LLM filters. 

\vspace{-1mm}
\paragraph{Reported tasks and dimensions}
For the evaluated dimensions reported in each paper, we apply \textit{PorterStemmer} from the \textit{nltk} library to normalize all dimension phrases.
We also group each paper's generation task (as reported by authors) into categories for analysis using the manually-curated word-stem mappings in Appendix~\ref{app:task_level_analysis}. 
The most frequent task categories are \textit{Dialogue and interactive systems} (n=40), \textit{Summarization} (n=26), \textit{Safety and jailbreaking} (n=26), \textit{QA} (n=18), and \textit{Story generation} (n=13). Evaluation dimensions show some consistency within task categories but are highly variable across studies; e.g., relevance is assessed across many task categories; coherence appears very often for story generation; correctness is most frequent in QA tasks. The most common dimensions per task and interpretation can be found in Appendix~\ref{app:task_level_analysis}.

\vspace{-1mm}
\paragraph{Form of human annotation tasks and reporting}
Human judgments are most commonly collected via \textit{binary} judgments (26\%) and \textit{pairwise comparison} (22\%), followed by \textit{likert scale} (22\%), \textit{numeric scale} (21\%), \textit{categorization} (13\%), and \textit{rank-based} (3\%) tasks.\footnote{These do not sum to 100\% since each paper may include more than one form of annotation.} Of annotated papers, 47\% include human evaluation details in both the main paper and appendix, while 33\% report this information only in the main paper and 20\% only in the appendix. 
\vspace{-2mm}
\subsection{Key observations}
\vspace{-1mm}
We report key findings below. Additional analysis over other collected variables can be found in Appendix~\ref{app:additional_analysis}.

\vspace{-1mm}
\paragraph{*CL papers pervasively under-report important aspects of human evaluation protocols.} 
As in Figure~\ref{fig:main}, while most papers report the evaluated dimensions (98\%), number of annotators (77\%), and number of annotated samples (85\%), all other criteria are reported far less frequently. Key information such as justification for the chosen evaluation dimensions and how the evaluation is described to annotators (task guidelines) is only present in around 50\% of papers. Important aspects of annotation design such as payment information (29\%) and IRB determination (11\%) are rarely reported. Very few papers report statistical metrics (9\%) and none use power analysis to derive sample sizes. Given the importance of empirical results in NLP research, the pervasive lack of statistical reporting is particularly troubling. Complete sample and bootstrap estimates can be found in Appendix~\ref{app:bootstrap-table}.

\begin{figure}[t!]
    \centering
    \includegraphics[width=\linewidth]{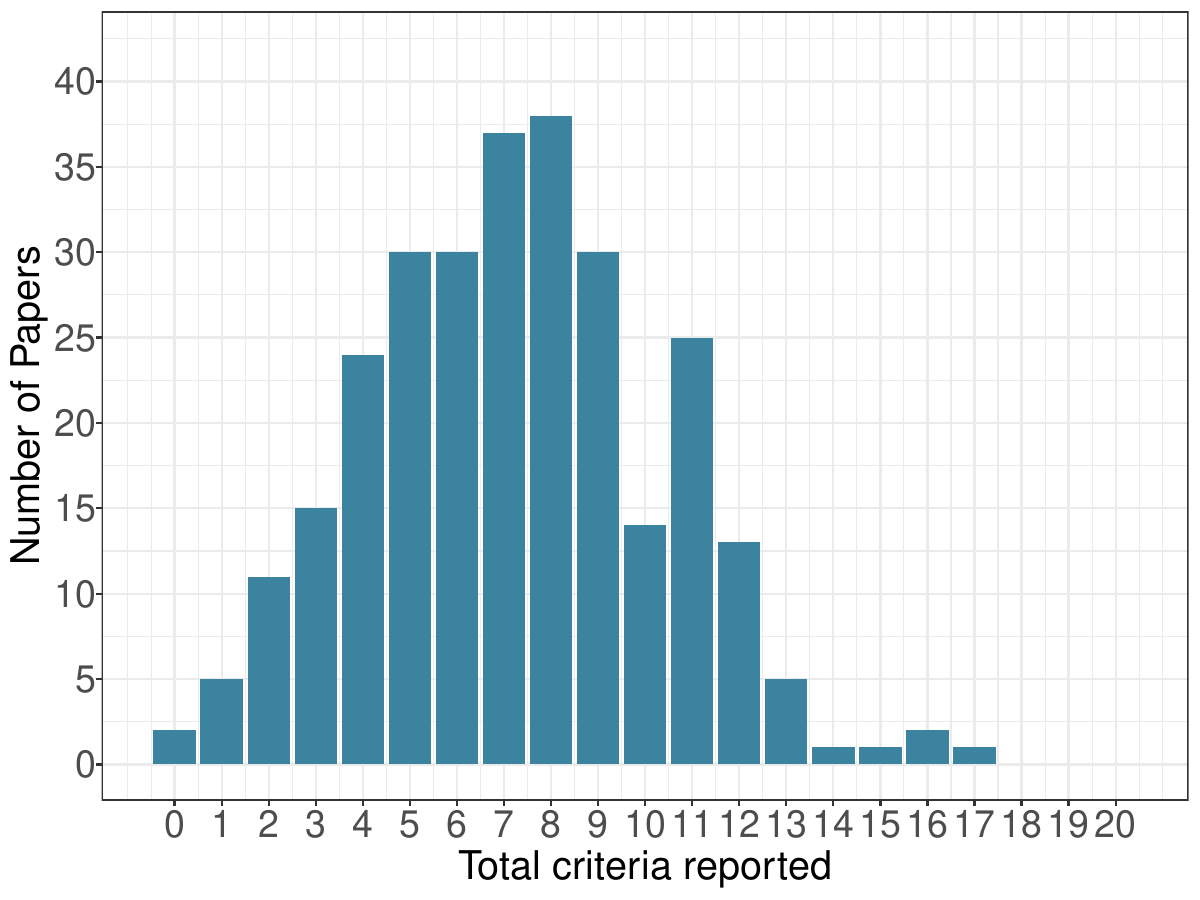}
    \vspace{-2mm}
    \caption{Distribution of total criteria reported; over half of papers report $\leq$7 of 20 reportable criteria. 
    }    
    \label{fig:distribution_of_total_reported_criteria}
\end{figure}

\vspace{-1mm}
\paragraph{Most papers report fewer than 8 criteria.} We find that the modal number of reported criteria is 7 out of a potential 20. More than half of papers we review report 7 or fewer reportable criteria (Median=7, SD=3)---note this is only whether an item is reported without any judgment on the content or sufficiency of what is reported. Very few papers (n=5) report more than 13 criteria, and no papers report all 20 reportable criteria.

\vspace{-1mm}
\paragraph{Norms around sample size and annotator count are strong but not well-justified.}
Annotation sample size and annotator count affect the inferences that can be drawn from a model evaluation study, yet this information is not consistently reported. We find that 15\% of papers do not report sample size, and 23\% do not report the number of annotators involved in human evaluation. Among papers that do report this information, we find no consistent practices for determining or justifying sample size (e.g., no papers use power analysis to determine sample size). Reported sample sizes vary widely, ranging from as few as 10 to a maximum of 23,040, with a median sample size of 170.  

\begin{figure}[t!]
    \centering
    \includegraphics[width=\linewidth]{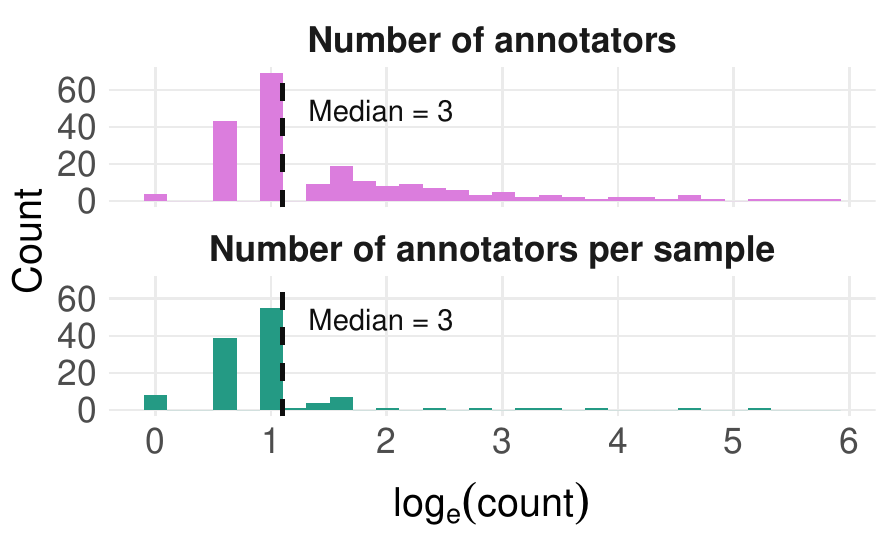}    \includegraphics[width=\linewidth]{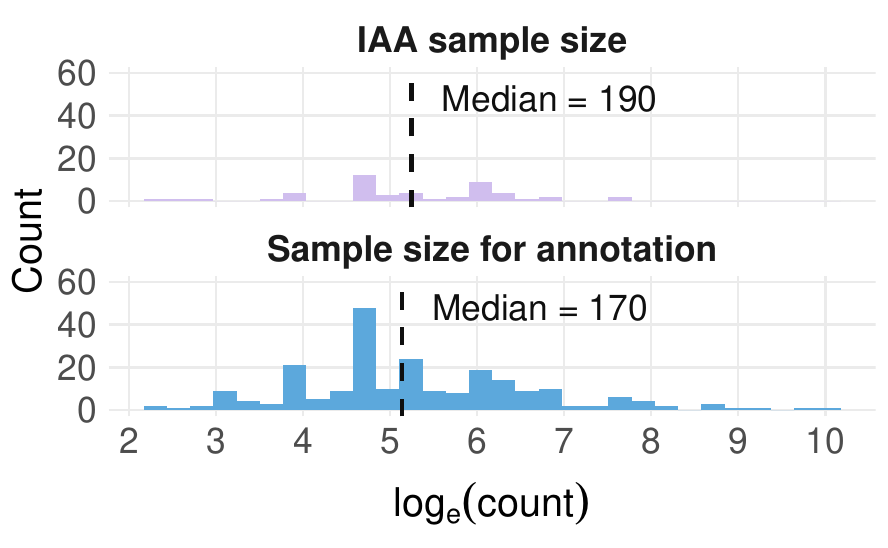}
    \vspace{-2mm}
    \caption{Distributions of annotator and sample counts across the manually annotated paper sample that reported relevant statistics in 2023-2025 *CL conferences.}    \label{fig:annotated_sample_annoator_distribution}
\end{figure}

Among papers that report the number of annotators (Median=3, MAD=1.48\footnote{MAD: median absolute deviation is a variability measure similar to standard deviation but less sensitive to outliers.}), three annotators is the most common configuration (32\%), followed by two annotators (20\%). Among papers reporting more than one annotator, only about half (51\%) report IAA, corresponding to 46\% of papers overall. Among papers that report IAA, the median number of samples used to compute agreement is 190. Figure~\ref{fig:annotated_sample_annoator_distribution} shows log-scale distributions. 
\add{While these statistics provide a useful high-level summary, we do not treat reporting IAA alone as sufficient evidence of evaluation quality. In some annotation settings, IAA may be inappropriate or less informative, such as when disagreements are resolved through consensus discussion. IAA metrics can also conflate different sources of disagreement, such as genuine subjectivity on the task versus problems in annotation design such as having unclear instructions or labels~\cite{fleisig-etal-2024-perspectivist}.}

\paragraph{Annotator information is usually missing or incomplete.} Despite increasing evidence of the influence of annotator background on annotation outcomes~\cite{sap2019risk,al-kuwatly-etal-2020-identifying,ding2022demographic}, *CL papers lack consistent reporting of annotator demographic information. We find that 29\% of papers do not report any demographic information about annotators, and 65\% do not report any information about recruitment platforms. Among papers that report some demographics, we find that 31\% of these papers recruit students, 50\% recruit domain experts (as described by authors), and 13\% of papers recruit paper authors as annotators. Among other characteristics, education is most frequently provided (48\%), followed by language (27\%), gender (12\%), and country of residence (9\%).

\begin{figure}[t!]
    \centering    \includegraphics[width=\linewidth]{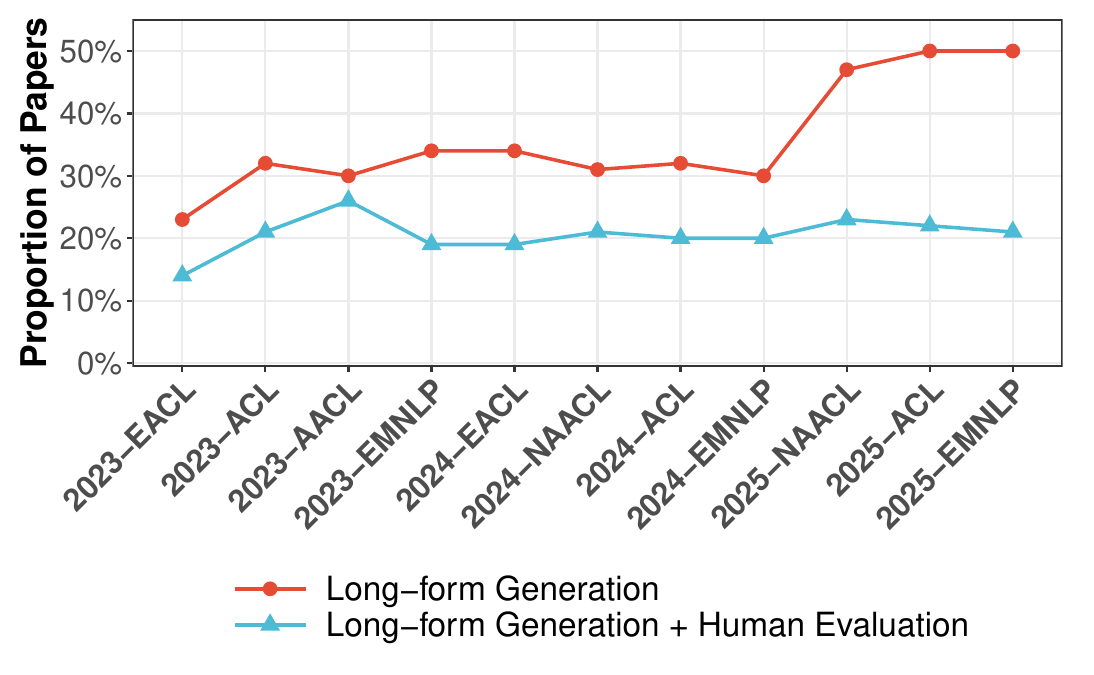}
    \vspace{-3mm}
    \caption{Temporal trends for 2023–2025 *CL conferences. While papers studying long-form generation have increased in the last year, the proportional use of human evaluation for these tasks has decreased.}
    \label{fig:humaneval_trend}
    \vspace{-4mm}
\end{figure}

\vspace{-2mm}
\paragraph{Annotation quality control is rarely employed.} We track whether researchers adopt any data filtering steps (i.e., attention checks, manipulation checks)---techniques to remove low-quality crowdsourced data or any other procedure to ensure annotation quality. We find that only 6\% of papers include data filtering steps, and 22\% of papers include procedures to ensure annotation quality. These procedures usually focus on ensuring annotation consistency, e.g., having a training or pre-assessment period for annotators, pilot studies to ensure annotation clarity, or methods to improve the reliability of annotation (such as only keeping samples with full agreement among annotators, or having multiple stages of quality checks). Around 30\% of papers report steps for resolving disagreements among annotations; majority vote (38\%) is the most common, followed by \textit{averaging} (24\%) and \textit{having a consensus process} (17\%). 
Only a small proportion (19\%) of papers discuss the limitations of their human evaluation pipeline.

\vspace{-1mm}
\subsection{Temporal trends}
\vspace{-1mm}
\paragraph{The proportion of long-form generation papers with human evaluation is declining.} While the number of studies of long-form generation tasks is increasing, the proportion using human evaluation is declining. As Figure~\ref{fig:humaneval_trend} and Table~\ref{tab:longform_proportion} show, while the proportion of papers studying long-form generation tasks has increased from around 30\% to 50\% in 2025, the proportion of overall papers that include human evaluation has been stable at around 20\%. We also observe that half of papers in our manually annotated set adopt LLM-judges for evaluation, and 20\% of these papers use LLM-judges for evaluation in other dimensions without human evaluation. 
\vspace{-1mm}
\paragraph{Use of human evaluation for meta-evaluation of LLM-judges is on the rise.} We observe the increasing use of human evaluation to assess the performance of LLM-judges (more in Appendix~\ref{app:llm-annotated analysis}). The proportion of papers using both LLM-judges and human evaluation for the same dimensions increased from 4\% in EMNLP '23 to 30\% for EMNLP'24 and 12\% for EMNLP'25. Meta-evaluation of LLM-judges also necessitates a reliable human evaluation pipeline; however, we did not identify obvious improvements in reporting among studies that use human evaluation for meta-evaluation in our manually annotated subset. 
\vspace{-2mm}

\section{Discussion \& Recommendations}
\label{sec:discussion}
\vspace{-1mm}
\label{sec:discussion}
Our analysis reveals clear gaps between the central role human evaluations play in NLP research, and the current rigor (or lack thereof) in reporting practices. Across *CL papers studying medium- and long-form generation from the last three years (2023--2025), we observe under-reporting of important criteria, high variability in human evaluation study design, and a recent and rapid shift in the way human judgments are used, especially in a meta-evaluative capacity for LLM judges. \add{We reiterate that our annotation is lenient, giving authors credit for reporting any information without judging the sufficiency or adequacy of the reported information. As a result, our estimates present a conservative picture of reporting gaps.} We discuss implications of our findings and outline recommendations for the community and for future work.

\vspace{-1mm}
\paragraph{R1: Report core reportable criteria for reproducibility.} 
While reporting details such as recruitment information, IAA, and task guidelines can take up important space, we argue that it is possible to report such crucial details succinctly.
For example, for the annotation study described in this paper, a passage such as the following provides most essential details: \\ [-3mm]

\footnotesize
\noindent We analyze reporting practices for human evaluation in *CL papers using a codebook of \totalquestions question, including \corequestions core reportable criteria associated with reproducible science. The codebook was iteratively developed based on the reproducible science framework, several rounds of pilot testing, iterative feedback from the research team, and prior work (cite). Using the codebook, we manually annotate \numannotated papers studying long-form generation with human evaluation drawn from the *CL corpus 2023--2025. Five annotators (2 PhD, 2 Masters, and 1 undergraduate student, all with experience reading and writing NLP papers), who are also authors of this paper, underwent a multi-week calibration process. All annotators met and exceeded an IAA threshold on a held-out set of 5 papers (155 questions per annotator), achieving 73\% agreement on binary questions ($\kappa$=0.54) and moderate agreement ($\kappa$=0.25-0.3) on multiple choice questions. For analysis, we report descriptive statistics with bootstrapped confidence intervals for reporting frequencies. Task guidelines and screenshots of our annotation interface are provided in the Appendix. \\ [-3mm]

\normalsize
\noindent We recommend at least this (based on our \corequestions criteria) as a minimum template for human evaluation reporting.

\vspace{-1mm}
\paragraph{R2: Tailor evaluation design to the needs of the study and document decisions} 
Papers evaluating the same task type (e.g., summarization, QA), can differ widely in evaluated dimensions (Appendix~Figure~\ref{fig:stemmed_dimension}), how those dimensions are operationalized, annotation formats, and analysis methods. We recognize that differences in study objectives can and should lead to differences in evaluation design. However, we recommend that researchers deliberately consider dimensions, scales, and evaluation protocols from prior work and whether to adopt or reject them. When new evaluation facets or decisions are introduced, rationales and operationalization details should be clearly described. Such justification is not always provided now (around half of papers provide some justification and only 25\% of papers justify dimensions using prior work). The current degree of heterogeneity can make it difficult to synthesize findings across studies or determine whether studies are measuring similar underlying constructs. 

\vspace{-1mm}
\paragraph{R3: Hold human evaluation to a higher standard} Temporal analysis highlights a shift in human evaluation practices, especially their increasing use for meta-evaluation of LLM-judges. 
If human evaluation is poorly documented or inconsistent across studies, it cannot serve as a reliable gold standard for assessing LLM-judges. Weak human evaluation protocols can introduce error into downstream systems, as shaky foundations lead to structural failure. Rather than reducing the importance of human evaluation, its growing meta-evaluative role demands increased rigor, transparency, and higher standards of documentation.

\vspace{-1mm}
\section*{Conclusion}
\vspace{-1mm}
Human evaluation remains a cornerstone of NLP research, especially for long-form and open-ended generation tasks. The  community could substantively improve its reporting practices with only modest changes in authoring norms as we have suggested above. We hope that the criteria list and recommendations presented here can serve as a practical reference point for the future evolution of human evaluation and documentation practices.

\section*{Limitations}
\vspace{-1mm}
Our meta-analysis is limited to papers in the past three years (2023--2025) and *CL conferences.  This leaves open questions about the reporting practices of NLP research papers in other conferences and journals, or adjacent research communities. We encourage future work to build on our motivation and methods to broaden the examination of evaluation and reporting practices in our research communities.

\add{We also acknowledge the limitations of the human annotation task introduced in this paper. Our annotation protocol actually consists of many distinct subtasks (e.g., extracting IAA, identifying the main NLP task studied by each paper), each of which presents unique challenges of information extraction and interpretation. We report IAA at the overall level of our annotation task, treating these subtasks as components of a whole, since item-level estimates would be too unstable to interpret reliably. This, however, masks variation in difficulty across the subtasks. Specifically, annotators found some subtasks to be more challenging due to larger variation in how and where information was presented, as well as in the amount of subjectivity needed for interpretation. We identified these variations and difficulties during annotator onboarding, and conducted random quality checks as a way to mitigate their impacts.}

In addition, we acknowledge that what is considered ``reportable'' can vary significantly depending on what task is being performed by models and assessed, and the role of the evaluation itself. Our work does not aim to critique any individual study for its design choices, but is geared towards understanding norms and patterns in the community as a whole and offering recommendations for how to improve documentation practices where there are clear gaps. We leave ample room for authors to interpret and justify what is or is not relevant for their evaluation setting. Nonetheless, it may be useful in future work to describe the needs specific to certain NLP tasks or user groups, perhaps through more granular or adaptive criteria lists, or based on evolving norms within those sub-communities.

While we propose lightweight reporting practices, we acknowledge that additional documentation requirements can introduce overhead. In fast-moving contexts, particularly those involving rapid model iteration, researchers may reasonably prioritize iteration speed over thorough reporting. However, when human evaluation is involved, we argue that a baseline level of documentation is necessary to support interpretation, reproducibility, and comparison of results. Adoption of our recommendations may vary depending on a research team's available resources and timelines, but tradeoffs should be made explicit rather than implicitly omitted from evaluation reporting.

\section*{Ethical considerations}
\vspace{-1mm}
We identify no immediate ethical concerns with our research study or conclusions. This study analyzes publicly available academic papers and does not involve collecting new data from human participants. All human annotation is conducted by the authors and trained collaborators on published materials, without collecting personal or sensitive information. As such, this work does not require institutional review board (IRB) review.

We acknowledge that human judgments may reflect annotator perspectives and subjective biases. To mitigate this, we focus on assessing whether elements are reported at all rather than judging the adequacy of what is reported. We support this by employing a structured codebook, annotator training, and by reporting our inter-annotator agreement. 

Our use of LLMs is limited to supporting large-scale analysis, and we recognize broader ethical concerns surrounding LLM-based evaluation, including bias propagation and over-reliance on automated judgment. Accordingly, we report human-annotated results as our primary findings, and use LLM annotations for supplementary evidence, applying a validation accuracy threshold of 0.75 to ensure reliability, as described previously.
\section*{Acknowledgments}
We thank Anthony Hevia for providing feedback on earlier versions of the annotation protocol. This work is supported by gifts from Google and the Allen Institute for AI. 

\bibliography{custom}

\appendix
\label{sec:appendix}

\section{Final codebook} \label{app:complete_criteria_list}
We include the complete criteria list and final codebook in Table~\ref{tab:reportable_criteria_full_list}. Specifically, we group questions into three categories: (i) task documentation, (ii) annotation design, and (iii) analysis \& interpretation. Starred items are included in our core criteria list of 20 reportable items. We include the exact questions that annotators answer when annotating each paper.

\begin{table*}[th!]
\setlength{\tabcolsep}{0pt}
\renewcommand{\arraystretch}{0.95}
\scriptsize
\centering
\caption{Full list of questions annotated for each paper. \ding{72} indicates those corresponding to core reportable criteria. }
\begin{tabular}{L{0.24cm} L{4.6cm} @{\hspace{3mm}} L{10.8cm}}
\toprule
& \textbf{Category \& Element Name} & \textbf{Question for Annotation} \\

\midrule
 & \textbf{Category: Task Documentation (5)} & \\
\cmidrule{2-3}
\ding{72} & Evaluated dimensions & What dimensions are annotators asked to evaluate regarding the models' output? \\
\ding{72} & Eval dimensions: justification provided & Is there justification provided for the selected dimensions or the human evaluation pipeline? \\
\ding{72} & Eval dimensions: justification by prior work & If yes to the previous question, is prior work cited? \\
\ding{72} & Task guidelines reported & Are task introductions/guidelines included? \\
\midrule
& \textbf{Category: Annotation Design (18)} & \\
\cmidrule{2-3}
 & Main Task & What is the main task the paper focuses on (e.g. summarization, dialogue)? \\ 
 &  & $\hookrightarrow$ If other for the last entry, provide a detailed description of the form of the task.\\ 
 & Domain & What is the domain that the main tasks related to (e.g.medicine, programming)? \\ 
 &  &  $\hookrightarrow$ If other is the response for the previous entry, describe the domain.\\ 
 
 & Longform generation & Free-form Generation Evaluation? \\ 
 & Human evaluation & Human evaluation? \\ 
 & More than one evaluation task & Is there more than one human evaluation pipeline? \\ 
 & Form of annotation task & What is the form of annotation task for human evaluation? \\ 
 &  & $\hookrightarrow$ If other for the last entry, provide a detailed description of the form of the task.\\ 
 & Claims task is novel & Do the authors claim the long-form generation task is newly introduced (novel)? \\ 
 & Sections w/ human eval details & Which section(s) include details about the human evaluation? \\ 
 & & $\hookrightarrow$ What is the location of the main design details of human evaluation (i.e.necessary information for reproducing the evaluation)? \\ 
 & Only human eval used &Is human evaluation the only evaluation method being used for assessing model performance? \\ 

 & LLMs used for eval & If no to the previous question, are LLMs being used to evaluate model outputs? \\ 
 & LLMs and humans eval same dimensions & If yes to the previous question, are human and LLMs evaluating the same dimensions of model outputs? \\ 
\ding{72} & Sample size for annotation & Total sample being annotated \\ 
\ding{72} & Power analysis used & Is sample size determined by power analysis? \\ 
\ding{72} & Recruitment platform & Recruitment platform (NA if not reported) \\ 
\ding{72} & Annotator inclusion/exclusion criteria reported & If recruitment platform is not NA, any restrictions on participation for annotation (Yes/No) \\ 
\ding{72} & Payment information reported & Is payment to annotator reported? \\ 
\ding{72} & IRB determination & Is IRB or similar ethics review process used? \\ 
\ding{72}  & Code/annotation interface reported & Is code for or image of the annotation interface shared?  \\
\ding{72} & Method for ensuring quality reported & Is there any procedure used to ensure human annotation quality (e.g., training period of annotators)? \\ 
&  & $\hookrightarrow$ If yes, please copy paste the exact text from the paper \\ 

\midrule
& \textbf{Analysis \& Interpretation (14)} & \\
\cmidrule{2-3}
\ding{72} & Annotator demographics & What demographic information of participants (if any) is reported?\\ 
 & Annotators are students & Are the human annotators students? \\ 
 & Annotators are authors & Are authors also annotators? \\ 
  & Annotators are experts & Are human annotators referred to as experts or have domain expertise? \\ 
  & & $\hookrightarrow$ What is the description of the annotators' expertise (copy and paste content from paper)? \\ 
\ding{72} & Number of annotators & Number of annotators \\ 
\ding{72} & Number of annotators per sample & Number of annotators for each annotated item \\ 
  
\ding{72} & IAA value reported & Is interrater agreement reported? \\ 
\ding{72} & IAA sample size & Number of samples used to compute IAA \\ 
  & IAA metrics & What metrics are reported for interrater agreement? \\ 
  & & $\hookrightarrow$ If other is selected for the previous question, write down the metric here.\\ 
\ding{72} & Data filtering steps reported & Are any filtering steps applied after human annotations are collected? (e.g., outlier removal, attention checks, manipulation checks) \\ 
 & Strength of agreement & How strong is the agreement (report agreement quality based on kappa interpretation)? \\ 
 & & $\hookrightarrow$ Comments on agreement description \\ 
\ding{72} & Disagreement resolution method & How is disagreement being treated? \\ 
\ding{72} & Statistical metric reported & Are any of the following metrics reported for the human evaluation data: standard error/deviation, confidence interval?\\ 
\ding{72} & Limitations discussed & Are there any limitations noted in regards to the human evaluation pipeline? \\ 
& & $\hookrightarrow$ Is yes to the previous column, record what authors mentioned regarding the limitation \\

\bottomrule
\end{tabular}
\label{tab:reportable_criteria_full_list}
\end{table*}

\section{Corpus construction: keyword filters}
\label{app:keyword_filters}
\begin{table*}[th!]
\centering
\small
\caption{Open-ended natural language generation keyword set used in \textbf{Step 1: Keyword filters}.}
\begin{tabularx}{\textwidth}{L{45mm}L{108mm}}
\toprule
\textbf{Task} & \textbf{Keywords Used for Filtering} \\
\midrule
General Long-form Keywords
& \texttt{long form}, \texttt{long-form}, 
\texttt{Summarisation/ Summarization} \\
\midrule
Summarisation / Summarization
& \texttt{Extractive Summarisation/ Summarization}, 
\texttt{Abstractive Summarisation/ Summarization}, 
\texttt{Multimodal Summarisation/ Summarization}, 
\texttt{Multilingual Summarisation/ Summarization}, 
\texttt{Conversational Summarisation/ Summarization}, 
\texttt{Query(-)focused Summarisation/ Summarization}, 
\texttt{Multi-document Summarisation/ Summarization}, 
\texttt{Multidocument Summarisation/ Summarization}, 
\texttt{Long(-)form Summarisation/ Summarization},
\texttt{Few(-)shot Summarisation/ Summarization}, 
\texttt{Document Summarisation/ Summarization}, 
\texttt{Text Summarisation/ Summarization}, 
\texttt{Opinion Summarisation/ Summarization}, \texttt{Review Summarisation/ Summarization}, 
\texttt{Legal Document Summarization}, 
\texttt{Scientific Paper Summarisation/ Summarization}, 
\texttt{News Summarisation/ Summarization}, 
\texttt{Explanatory Summarisation/ Summarization} \\
\midrule
Narrative \& Story Generation
& \texttt{Narrative Generation}, \texttt{Story Generation} \\
\midrule
Question Answering
& \texttt{Long-Form Question Answering}, 
\texttt{Long Form Question Answering}, 
\texttt{Open-Domain Question Answering}, 
\texttt{Open Domain Question Answering}, 
\texttt{Explanatory Question Answering}, 
\texttt{Document-based Question Answering}, 
\texttt{Document Question Answering},
\texttt{Long-Form QA}, \texttt{Long Form QA}, 
\texttt{Open-Domain QA}, 
\texttt{Open Domain Question Answering}, 
\texttt{Explanatory QA}, 
\texttt{Document-based QA}, 
\texttt{Document QA} \\
\midrule
Conversational Systems
& \texttt{Reading Comprehension}, 
\texttt{Dialogue}, \texttt{Dialog}, \texttt{Conversation}, 
\texttt{Conversational AI}, \texttt{Dialogue Management}, 
\texttt{Conversational Agent}, \texttt{Chatbot}, 
\texttt{Conversational Interface}, \texttt{Dialogue System}, 
\texttt{Chat-oriented Dialogue System}, 
\texttt{Chat oriented Dialogue System}, 
\texttt{Open-domain Conversational System}, 
\texttt{Open domain Conversational System}, 
\texttt{Closed-domain Conversational System}, 
\texttt{Closed domain Conversational System} \\
\midrule
Report \& Writing Generation
& \texttt{Report Generation}, \texttt{Essay Generation}, 
\texttt{Script Writing}, \texttt{Book Writing}, 
\texttt{Content Creation}, 
\texttt{Extended Abstract Generation}, 
\texttt{Technical Documentation Generation}, 
\texttt{Healthcare Documentation}, 
\texttt{Collaborative writing}, 
\texttt{open-ended generation} \\
\midrule
Editing \& Research
& \texttt{deep research}, 
\texttt{text simplification}, 
\texttt{paraphrasing}, 
\texttt{document editing} \\

\bottomrule
\end{tabularx}

\label{tab:exact_longform_task_keyword_mapping}
\end{table*}

In Table~\ref{tab:exact_longform_task_keyword_mapping}, we provide the complete set of keywords used to identify papers studying long-form text generation. Keywords are matched in a case-insensitive manner with stemming against titles, abstracts, and main text extracted using GROBID. Papers matching at least one keyword are retained for subsequent LLM-based filtering.

\section{Corpus construction: LLM filters}
\label{app:llm_filters}

\begin{figure*}[th!]
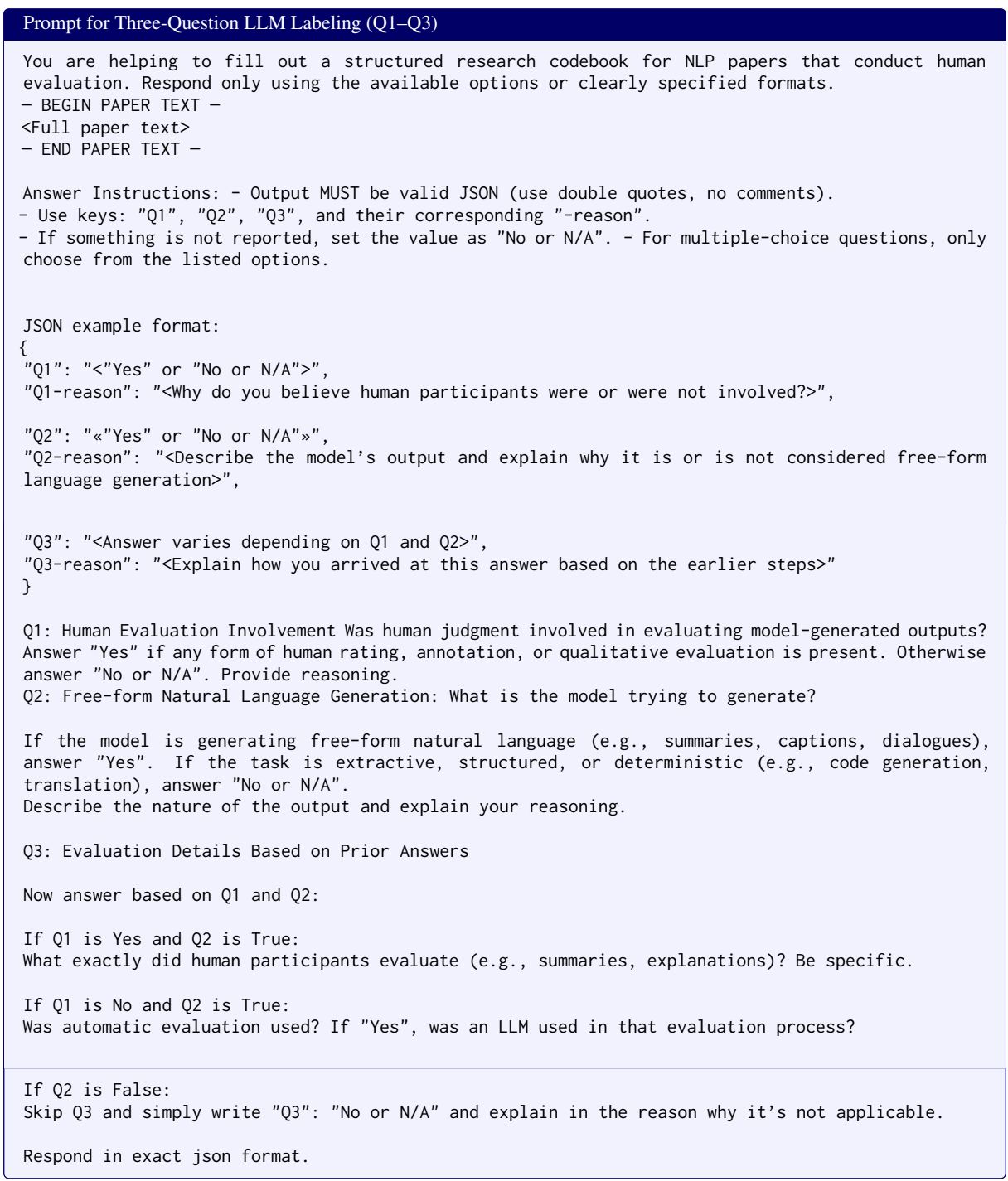

\centering
\small
\begin{tcolorbox}[
    promptbox,
    width=\textwidth,
    title=Prompt for Three-Question LLM Labeling (Q1--Q3),
    breakable,
    enhanced,
    boxrule=0.5pt,
    left=2mm,
    right=2mm,
    top=1mm,
    bottom=1mm
]
\ttfamily
You are helping to fill out a structured research codebook for NLP papers that conduct human evaluation.
Respond only using the available options or clearly specified formats.

--- BEGIN PAPER TEXT ---\\
<Full paper text>\\
--- END PAPER TEXT ---\\

Answer Instructions:
- Output MUST be valid JSON (use double quotes, no comments).\\
- Use keys: "Q1", "Q2", "Q3", and their corresponding "-reason".\\
- If something is not reported, set the value as "No or N/A".
- For multiple-choice questions, only choose from the listed options.\\
\\

JSON example format:\\
\{\\
    "Q1": "<"Yes" or "No or N/A">",\\
    "Q1-reason": "<Why do you believe human participants were or were not involved?>",\\

    "Q2": "<<"Yes" or "No or N/A">>",\\
    "Q2-reason": "<Describe the model's output and explain why it is or is not considered free-form language generation>",\\

    "Q3": "<Answer varies depending on Q1 and Q2>",\\
    "Q3-reason": "<Explain how you arrived at this answer based on the earlier steps>"\\
    \}\\

Q1: Human Evaluation Involvement
Was human judgment involved in evaluating model-generated outputs?
Answer "Yes" if any form of human rating, annotation, or qualitative evaluation is present.
Otherwise answer "No or N/A". Provide reasoning.

Q2: Free-form Natural Language Generation: What is the model trying to generate?\\

    If the model is generating free-form natural language (e.g., summaries, captions, dialogues), answer "Yes".
    If the task is extractive, structured, or deterministic (e.g., code generation, translation), answer "No or N/A".\\
    Describe the nature of the output and explain your reasoning.\\

Q3: Evaluation Details Based on Prior Answers\\

Now answer based on Q1 and Q2:\\

    If Q1 is Yes and Q2 is True:\\
    What exactly did human participants evaluate (e.g., summaries, explanations)? Be specific.\\

    If Q1 is No and Q2 is True:\\
    Was automatic evaluation used? If "Yes", was an LLM used in that evaluation process?\\

    If Q2 is False:\\
    Skip Q3 and simply write "Q3": "No or N/A" and explain in the reason why it's not applicable.\\

    Respond in exact json format.

\end{tcolorbox}
\caption{Prompt used for LLM-based filtering to identify papers studying long-form generation tasks and which employ human evaluation. Papers satisfying both conditions are included for manual annotation (through stratified sampling) and LLM-assisted annotation.}
\label{prompt:prompt_three_question}
\end{figure*}

The full prompt text used for LLM-based second-stage corpus filtering is reproduced in Figure~\ref{prompt:prompt_three_question}.

\begin{table*}[th!]
\centering
\small
\caption{Keyword list used to identify and extract human evaluation sections from papers.}
\label{tab:human_eval_section_filter_keywords}
\begin{tabularx}{\linewidth}{L{45mm}L{108mm}}
\toprule
\textbf{Category} & \textbf{Human Evaluation Section Selection Keywords} \\
\midrule

Human Evaluation Indicators
& \texttt{human evaluation}, \texttt{manual evaluation}, \texttt{expert evaluation}, 
\texttt{human judg}, \texttt{human assess}, \texttt{expert assess}, 
\texttt{human preference}, \texttt{expert preference}, 
\texttt{user study}, \texttt{human study}, \texttt{participant}, \texttt{annotator}, \texttt{rater}, \texttt{subject}, 
\texttt{evaluator}, \texttt{human subject}, \texttt{human judgment},\texttt{interface}, \texttt{screenshot} \\

Evaluation Setup \& Protocol
& \texttt{Likert}, \texttt{pairwise}, \texttt{A/B}, \texttt{MOS}, \texttt{rating}, \texttt{assessment}, \texttt{preference}, \texttt{inter-annotator}, \texttt{Cohen's kappa}, \texttt{Krippendorff} \\

Recruitment Platforms \& Payment
& \texttt{AMT}, \texttt{Mechanical Turk}, \texttt{mturk}, \texttt{Prolific}, \texttt{crowdsourcing}, \texttt{paid}, \texttt{volunteer}, \texttt{Upwork},\texttt{IRB}, \texttt{consent}, \texttt{compensation} \\
\bottomrule
\end{tabularx}
\end{table*}

\section{Details for manual annotation}
\subsection{Annotation codebook reference} \label{app:codebook_reference}

\begin{table*}[th!]
\footnotesize
\caption{Codebook Reference Sheet: these clarifications of codes and answer options are provided to annotators.}
\centering
\renewcommand{\arraystretch}{1.05}
\begin{tabular}{L{3.6cm} L{2cm} L{9.5cm}}
\toprule
\textbf{Annotation field} & \textbf{Options} & \textbf{Clarification} \\
\midrule
\multirow{2}{=}{Are task introductions or guidelines for human evaluation included?} & Yes 
& Paper describes task introduction and instructions for annotators. \\
& No or N/A &  \\ 
& \\
\midrule
\multirow{6}{=}{What is the \textbf{domain} that the main task is related to (e.g., medicine, programming)?} 
& General & Select \textit{General} if no specific domain is related. \\
& Medicine & \\
& Legal & \\
& Coding/Programming & \\
& Journalism & \\
& Other &\\
\midrule
\multirow{2}{=}{Long-form Generation Evaluation?} 
& Yes 
& Free-form natural language (e.g., summaries, captions, dialogues); answer ``Yes''. \\
& No or N/A 
& If the task is extractive, structured, or deterministic (e.g., code generation, translation), answer ``No or N/A''. \\
\midrule
\multirow{2}{=}{Human evaluation?} 
& Yes 
& If the study involves human participants for evaluation of model-generated outputs (e.g., including benchmark papers where humans assess LLM-generated outputs to curate a benchmark; \textbf{exclude} benchmark papers if humans are only used to provide data). \\
& No or N/A 
&  \\
\midrule
\multirow{2}{=}{Is there more than one human evaluation pipeline?} 
& Yes 
& Yes if there are human evaluations used for separate tasks or procedures in the study. \\
& No or N/A 
&  \\
\midrule
Which \textbf{section(s)} include details about the human evaluation? [comma-separated list] 
& Open-ended 
& Copy and paste the section name(s) which involve details of the human evaluation. \\
\midrule
\multirow{4}{=}{What is the \textbf{location} of the main design details of human evaluation (i.e., necessary information for reproducing the evaluation)?} 
& main 
& Select this option if the main details about the human evaluation pipeline (e.g., recruitment, task description, samples) are included in the main paper. \\
& appendix 
& Select this option if the main details about the human evaluation pipeline (e.g., recruitment, task description, samples) are included in the appendix. \\
& both 
& Select this option if the main details about the human evaluation pipeline (e.g., recruitment, task description, samples) are included in both the main paper and the appendix. \\
& neither 
& Select this option if any information related to human evaluation is not found anywhere. \\
\midrule
\multirow{2}{=}{Is human evaluation the only evaluation method being used for assessing model performance?} 
& Yes 
& This means there are no automatic metrics and no LLMs used to evaluate model performance. \textbf{Only human participants} are used to evaluate model outputs. \\
& No or N/A 
&  \\
\midrule
Total sample being annotated & &Count unique examples presented for human annotation. \\ 
\midrule
\multirow{2}{=}{Is \textbf{interrater agreement} reported?} & Yes 
& If the study mentions interrater agreement among annotators. \\
& No or N/A &  \\ 
\midrule
\multirow{2}{=}{Is the number of samples used to compute IAA reported?} & Yes & Yes only if the authors describe exact sample counts or state all annotators annotated all samples. \\ 
& No or N/A &Often not explicitly mentioned.\\
\midrule
\multirow{2}{=}{How strong is the agreement?} & [Select Options]&Interpret numeric values using standard kappa guidelines if no interpretation is provided.\\ 
& NA &if no agreement is reported. \\ 
\midrule
\multirow{5}{=}{How is disagreement treated?} & Majority vote 
&  \\
& Average & \\
& Pick one &  \\
& Consensus process is applied & A group decision-making process aiming for broad agreement. \\
& Other &  \\
\bottomrule
\end{tabular}
\label{tab:codebook_reference-sheet}
\end{table*}
Instructions for annotation task and item answers are reproduced in Table~\ref{tab:codebook_reference-sheet}. For criteria that are difficult to assess, we clarify each answer option to maximize annotation consistency.

\subsection{Annotation interface}
\label{app:annotation_interface}
\begin{figure*}[th!]
    \centering
\includegraphics[width=\linewidth]{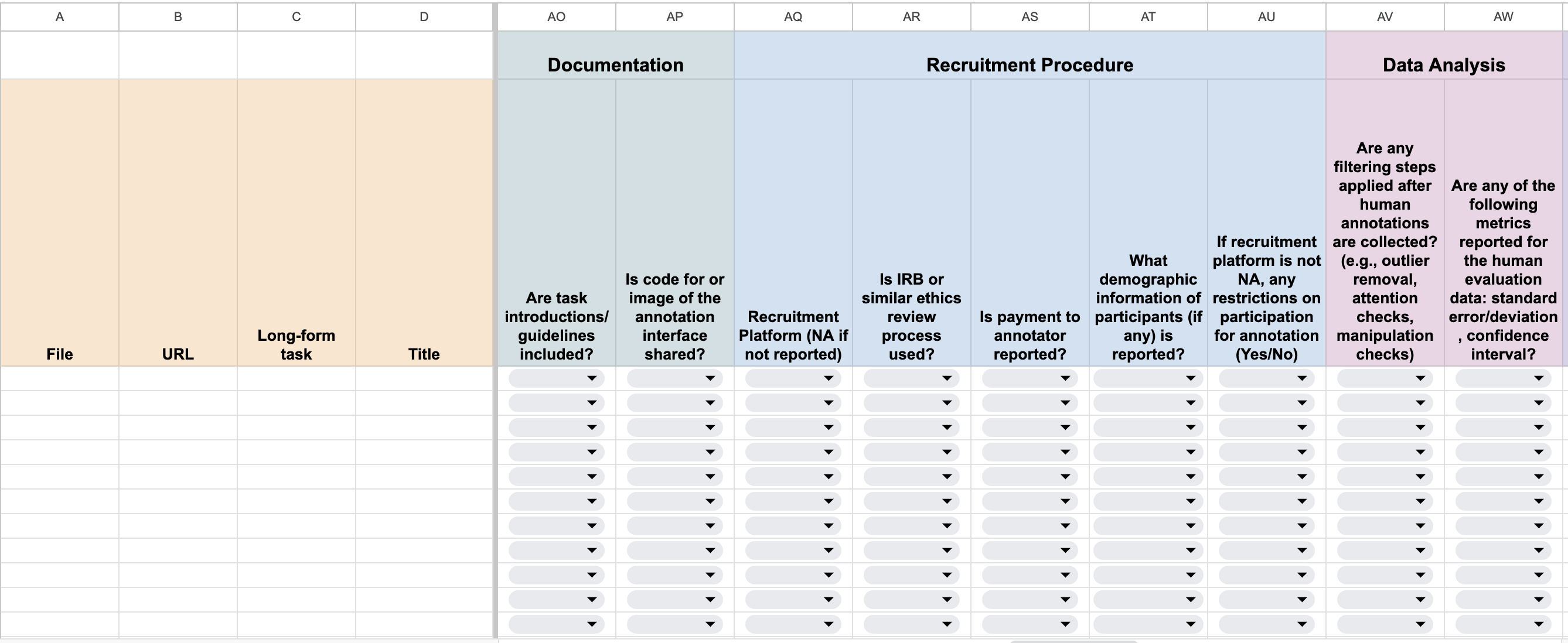}
    \caption{Partial screenshot of our annotation interface in Google Sheets showing questions pertaining to documentation, recruitment procedure, and data analysis. These are 9 out of the total 37 questions that annotators answer for each paper they annotate.}
\label{fig:annotation_interface}
\end{figure*}

In Figure~\ref{fig:annotation_interface}, we include a partial screenshot of the annotation interface we develop in Google Sheets. Answer options are restricted to valid types.

\section{Details for LLM-assisted annotation}
\label{app:llm-annotation-details}

\subsection{Keywords for section selection}
\label{app:human-eval-section-keywords}

Table~\ref{tab:human_eval_section_filter_keywords} provides the complete list of keywords and phrases used to identify human evaluation sections in each paper. Keywords are matched in a case-insensitive manner with stemming and are used to select candidate sections, which are then passed to the LLM-based annotation prompt (see Figure~\ref{prompt:question_intro}).
\subsection{LLM selection \& validation}
\label{app:llm_selection}
We conduct a pilot study to select an appropriate large language model for automatic annotation of human evaluation details. We compared three state-of-the-art models: Gemini-2.5-Pro, Claude-3.7-Sonnet-20250219, and GPT-4o-mini-2025-04-16 with identical prompts and input contexts. Performance is validated on the manual annotations of a held-out set consisting of 26 papers. For each model, we assess annotation quality using question-level validation accuracy, measuring consistency with human-annotated ground truth across the full set of codebook questions. 

Among the models we test, GPT-4o-mini-2025-04-16 achieves the highest overall accuracy. Based on this empirical comparison, we select GPT-4o-mini-2025-04-16 as the annotation model for the remainder of our corpus.

\subsection{Prompts for LLM-assisted annotation}
\label{app:llm_annotation_prompts}

To control context length and improve reliability, we split codebook questions into five semantically coherent chunks for prompting. Questions from each chunk are answered in separate API calls, with the model instructed to return a flat JSON object with answers.

Prompts used for LLM-assisted annotation are reproduced in three figures: (i) the task introduction in Figure~\ref{prompt:question_intro}, (ii) the chunked question structure in Figure~\ref{prompt:question_chunk}, and (iii) the full list of annotation questions in Figure~\ref{prompt:question_list}. We validate LLM-based annotation using GPT-4o on a held-out set of 125 manually-annotated papers (25 from each of the five annotators), or 3,875 annotations in total (31 binary or multiple choice questions for each of the 125 papers). 
Table~\ref{tab:llm_validation_accuracy} reports the percentage agreement between GPT-4o and human annotations.

\newfloat{promptchunk}{tbp}{lop}
\floatname{promptchunk}{Chunk Prompt}

\newfloat{promptlist}{tbp}{lop}
\floatname{promptlist}{Question List}

\newfloat{promptintro}{tbp}{lop}
\floatname{promptintro}{Prompt Intro}

\begin{figure*}[th!]
\centering
\begin{tcolorbox}[promptbox, width=\textwidth, title={Prompt Intro}, label={prompt:question_intro}]

\footnotesize
You are an expert NLP paper auditor.

You will read three PAPER CONTENT blocks that precede this instruction in the prompt:
\begin{itemize}[leftmargin=*]
  \item ABSTRACT
  \item INTRODUCTION
  \item HUMAN-EVAL FILTERED (sections labeled MAIN/APPENDIX)
\end{itemize}

Answer only using evidence from those blocks. Do not infer beyond them.  
Be deterministic and conservative: if you are unsure, answer ``NA'' or ``No or N/A''.

\medskip
\textbf{Strict output rules (per chunk)}
\begin{itemize}[leftmargin=*]
  \item Answer the questions based on the \textbf{first human evaluation pipeline} mentioned in the paper.
  \item \textbf{Schema lock (per chunk):} output only the keys for this chunk and their corresponding
        reason fields (e.g., \texttt{Q8}, \texttt{Q8\_reason}).  
        Do not include any other \texttt{Q*} keys.
  \item Use labels verbatim where specified (e.g., ``Yes'', ``No or N/A'', ``main'', ``appendix'', ``both'', ``neither'').
  \item For numeric answers, return only numerals (e.g., ``3'').
  \item For list items, return a single comma-separated string.
  \item For each answer, also output the corresponding \texttt{Q*\_reason} with a short quote or paraphrase  
        (maximum 40 words, include section/page pointers if available).
  \item Output exactly one flat JSON object (no prose, no code fences).
\end{itemize}
\end{tcolorbox}
\caption{Prompt for LLM-assisted annotation: input prompt structure for each LLM call.}
\label{prompt:question_intro}
\end{figure*}

\begin{figure*}[th!]
\centering
\begin{tcolorbox}[promptbox, width=\textwidth,
                  title={Chunk Structure (Q1--Q46)}]
\small
Chunk 1 (Q1--Q7): Overview \& Task Setup \\
Chunk 2 (Q8--Q18): Human Evaluation – Overview \& Design \\
Chunk 3 (Q19--Q33): Human Evaluation – Task 1 Details (Annotators, Samples, IAA) \\
Chunk 4 (Q34--Q40): Documentation \& Recruitment \\
Chunk 5 (Q41--Q46): Data Analysis, Quality, and Limitations
\end{tcolorbox}
\caption{Prompt for LLM-assisted annotation: chunk structure for codebook questions.}
\label{prompt:question_chunk}
\end{figure*}


\begin{figure*}[th!]
\centering
\begin{tcolorbox}[promptbox, width=\textwidth, title={Full Question List}, label={prompt:question_list}]

\scriptsize

Q1: Free text (central empirical task, e.g., summarization, dialogue, QA, information extraction, data-to-text, evaluation/benchmarking, classification).

Q2: ACL tracks. ALWAYS answer "NA".

Q3: What is the domain of the main task? options: General, Medicine, Legal, Coding/Programming, Finance. If none is specific, answer "General".

Q4: "Yes" if free-form natural language generation; otherwise "No or N/A".

Q5: "Yes" if humans evaluate model-generated outputs (exclude benchmarks where humans only supply dataset labels); otherwise "No or N/A".

Q6: "Yes" if there is more than one human evaluation pipeline included in the paper; otherwise "No or N/A".

Q7: "Yes" if authors claim they proposed a novel NLP task; otherwise "No or N/A".

Q8: Comma-separated section numbers \& names that include human-eval details.

Q9: What is the location of the main design details of human evaluation (i.e. necessary information for reproducing the evaluation); options: "main" / "appendix" / "both" / "neither".

Q10: "Yes" if human evaluation is the ONLY evaluation used to assess model performance; otherwise "No or N/A".

Q11: "Yes" if LLMs are used to evaluate model-generated outputs; otherwise "No or N/A".

Q12: "Yes" if humans and LLMs evaluate the SAME dimensions; otherwise "No or N/A".

Q13: Free text (e.g., coherence, human-like, appropriateness) as a comma-separated list.

Q14: "Yes" if justification for human evaluation dimensions selection is provided; otherwise "No or N/A".

Q15: "Yes" if prior work is cited for justification of human evaluation dimensions; otherwise "No or N/A".

Q16: "Yes" if prior work is cited for the pipeline used in human evaluation; otherwise "No or N/A".

Q17: What is the form of annotation task for human evaluation?  
Choose one or more: binary, user studies, numeric scale, pair-wise comparison, likert scale, rank-based, categorization, other.

Decide using these rules (don’t rely on numbers alone):\\
• likert scale: Discrete ordinal options with verbal anchors (e.g., strongly disagree…strongly agree; poor/fair/good/very good/excellent; very bad…very good). Numbers (1–5/7) may appear but anchors define the scale. Keywords: “Likert(-type)”, agree/disagree, poor/good/excellent, very/slightly.\\
• numeric scale: Pure numeric ratings without Likert-style anchors, often MOS/continuous (e.g., MOS 0–100, “give a score from 1–10” with no named categories). Keywords: “MOS”, “Mean Opinion Score”, “0–100”, “points” with no anchors.\\
• pair-wise comparison: A vs B preference.\\
• rank-based: Order multiple systems/items (best→worst, top-k).\\
• binary: Yes/No, Correct/Incorrect, Accept/Reject.\\
• categorization: Choose a category label (e.g., error type A/B/C).\\
• user studies: Interactive/usability tasks with the system (e.g., SUS/UX), not isolated output judgements.

Priority / defaults:\\
1) If any verbal anchors are present (even alongside numbers)\\
2) If explicitly MOS or purely numeric with no anchors: numeric scale.\\
3) If both terms appear, prefer likert scale due to anchors.\\
4) If ambiguous “rate 1–5 quality” and anchors are implied or unclear: choose likert scale.\\
Output for Q17 must be a comma-separated subset of: binary, user studies, numeric scale, pair-wise comparison, likert scale, rank-based, categorization, other.

Q18: If "other" in Q17, describe (free text); else "NA".

Q19: Number of annotators. options: Numeric or "No or N/A".

Q20: "Yes" if annotators are referred to as experts or have domain expertise; otherwise "No or N/A".

Q21: Copy/paste expertise description.

Q22: "Yes" if annotators are students; otherwise "No or N/A".

Q23: "Yes" if authors are annotators; otherwise "No or N/A".

Q24: Total unique examples annotated (numeric) or "No or N/A".

Q25: Annotators per item (numeric) or "No or N/A".

Q26: "Yes" if power analysis determines sample size; otherwise "No or N/A".

Q27: "Yes" if IAA is reported; otherwise "No or N/A".

Q28: "Yes" if \#samples for IAA is reported; otherwise "No or N/A".

Q29: Numeric \#samples for IAA (if reported) or "No or N/A".

Q30: What metrics are reported for interrater agreement? options: Cohen's kappa, Fleiss' kappa, Krippendorff's alpha, Percent agreement, Pearson, Kendall tau, Intraclass Correlation Coefficient, Other, NA.

Q31: How strong is the agreement (report the average agreement level based on kappa interpretation) options: <0 No Agreement, 0–0.20 Slight, 0.21–0.40 Fair, 0.41–0.60 Moderate, 0.61–0.80 Substantial, 0.81–1.00 Almost perfect, or "NA" if not reported.

Q32: Free text or "NA".

Q33: How is disagreement being treated? options: "Majority vote" / "Average" / "Pick One" / "Consensus process is applied" / "Other" / "NA".

Q34: "Yes" if full text of task introductions/guidelines of human evaluation are included; otherwise "No or N/A".

Q35: "Yes" if code or image of the interface is shared; otherwise "No or N/A".

Q36: Recruitment Platform (NA if not reported): Volunteers / Upwork / Prolific / AmazonTurk / Other / NA.

Q37: "Yes" if IRB/ethics review is used; otherwise "No or N/A".

Q38: "Yes" if payment to annotators is reported; otherwise "No or N/A".

Q39: What demographic information of participants (if any) is reported? options: education, age, gender, language, Residence country/Location, other, No or N/A.

Q40: "Yes" if participation restrictions exist when recruiting annotators via platform; otherwise "NA".

Q41: "Yes" if post-annotation filtering (outliers, attention/manipulation checks); otherwise "NA".

Q42: "Yes" if SE/SD/CI reported; otherwise "No or N/A".

Q43: "Yes" if procedures ensure annotation quality (e.g., training); otherwise "No or N/A".

Q44: If Q43 is "Yes", exact text (free text); else "No or N/A".

Q45: "Yes" if limitations of human-eval pipeline are noted; otherwise "No or N/A".

Q46: If Q45 is "Yes", copy/paste limitation text (free text); otherwise "No or N/A".
\end{tcolorbox}
\caption{Prompt for LLM-assisted annotation: full question schema used for LLM annotation.}
\label{prompt:question_list}
\end{figure*}

\begin{table*}[h!]
\small
\centering
\renewcommand{\arraystretch}{1.15}
\caption{GPT-4o validation accuracy on held-out set of 125 papers. Only questions with validation accuracy greater than 0.75 (shown in \textbf{bold}) meet our criteria for presenting results.}
\begin{tabular}{L{8.5cm} L{3cm} L{2.5cm}}
\toprule
\textbf{Question} & \textbf{Type} & \textbf{Validation Acc.} \\
\midrule

\textbf{\underline{Category: Task Documentation}} & & \\
Eval dimensions: justification provided 
& Binary 
& 0.52 \\
\textbf{Eval dimensions: justification by prior work} 
& Binary 
& \textbf{0.81} \\
Task guidelines reported 
& Binary 
& 0.50 \\

\midrule
\textbf{\underline{Category: Annotation Design}} & & \\
Domain 
& Multiple choice 
& 0.72 \\
{Longform generation} 
& Binary 
& {0.66} \\
\textbf{Human evaluation} 
& Binary 
& \textbf{0.78} \\
More than one evaluation task 
& Binary 
& 0.68 \\
Form of annotation task 
& Multiple choice 
& 0.34 \\
Claims task is novel 
& Binary 
& 0.54 \\
Sections w/ human eval details
& Multiple choice 
& 0.45 \\
\textbf{Only human eval used} 
& Binary 
& \textbf{0.91} \\
LLMs used for eval 
& Binary 
& 0.72 \\
\textbf{LLMs and humans eval same dimensions} 
& Binary 
& \textbf{0.76} \\
\textbf{Power analysis used} 
& Binary 
& \textbf{1.00} \\
Recruitment platform 
& Multiple choice 
& 0.71 \\
\textbf{Annotator inclusion/exclusion criteria reported} 
& Binary 
& \textbf{0.83} \\
\textbf{Payment information reported} 
& Binary 
& \textbf{0.78} \\
\textbf{IRB determination} 
& Binary 
& \textbf{0.88} \\
Code/annotation interface reported 
& Binary 
& 0.71 \\
Method for ensuring quality reported 
& Binary 
& 0.58 \\
\midrule

\textbf{\underline{Category: Analysis \& Interpretation}} & & \\
Annotator demographics 
& Multi-label 
& 0.58 \\
\textbf{Annotators are students} 
& Binary 
& \textbf{0.84} \\
\textbf{Annotators are authors} 
& Binary 
& \textbf{0.86} \\
Annotators are experts 
& Binary 
& 0.74 \\
\textbf{IAA value reported} 
& Binary 
& \textbf{0.78} \\
IAA metrics 
& Multi-label 
& 0.54 \\

\textbf{Data filtering steps reported} 
& Binary 
& \textbf{0.90} \\
{Strength of agreement}
& Multiple choice 
& 0.58 \\
Disagreement resolution method 
& Multiple choice 
& 0.74 \\
\textbf{Statistical metric reported (SE/SD/CI)} 
& Binary 
& \textbf{0.94} \\
{Limitations discussed} 
& Binary 
& {0.58} \\

\bottomrule

\end{tabular}
\label{tab:llm_validation_accuracy}
\end{table*}

\section{Task-level analysis}
\label{app:task_level_analysis}
\begin{table*}[th!]
\centering
\small
\caption{Keyword stem-to-category mapping used to assign papers to primary NLP tasks for visualization.}
\label{tab:task_category_mapping}
\begin{tabularx}{0.85\linewidth}{L{50mm}L{80mm}}
\toprule
\textbf{Category} & \textbf{Stem Keywords Used for Mapping} \\
\midrule
Dialogue \& Interactive Systems 
& \texttt{dialog}, \texttt{dialogu}, \texttt{convers}, \texttt{interact}, \texttt{empathi} \\

Summarization 
& \texttt{summar}, \texttt{summari}, \texttt{summarizast} \\

Question Answering 
& \texttt{question}, \texttt{qa}, \texttt{answer} \\

Safety \& Jailbreak 
& \texttt{safeti}, \texttt{align}, \texttt{harm}, \texttt{jailbreak}, \texttt{hallucin}, 
\texttt{toxic}, \texttt{hate}, \texttt{privaci}, \texttt{inappropri} \\

Reasoning \& Planning 
& \texttt{reason}, \texttt{plan}, \texttt{logic}, \texttt{multihop}, \texttt{deduct}, 
\texttt{induct}, \texttt{counterfactu}, \texttt{think} \\

Instruction \& Prompting 
& \texttt{instruct}, \texttt{prompt} \\

Story Generation 
& \texttt{stori}, \texttt{narr}, \texttt{novel}, \texttt{drama} \\

Style Transfer
& \texttt{style}, \texttt{simplif} \\

Misinformation Detection
& \texttt{fake}, \texttt{misinform}, \texttt{fallaci} \\

Caption Generation
& \texttt{caption}, \texttt{script} \\

Personalized Generation
& \texttt{persona}, \texttt{person}, \texttt{role} \\

Information Retrieval
& \texttt{extract}, \texttt{inform}, \texttt{retriev} \\

\bottomrule
\end{tabularx}
\end{table*}

\paragraph{Task-category stem-keyword mapping} Table~\ref{tab:task_category_mapping} shows the stem keyword–to–task-category mapping used to assign each paper to a primary NLP task for task-level analysis.

\begin{figure*}[th!]
    \centering    \includegraphics[width=1\linewidth]{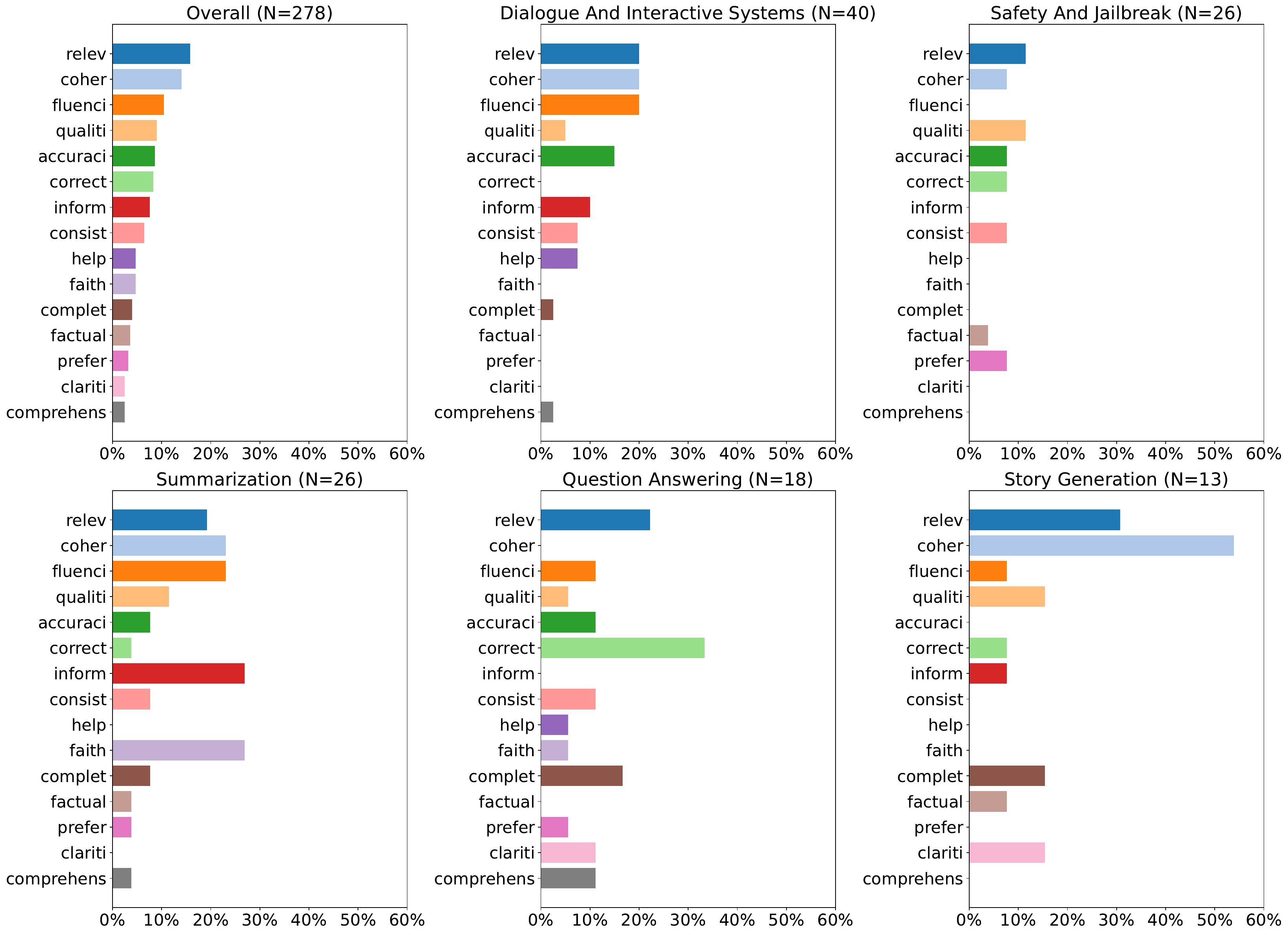}
    \caption{Distribution of stemmed evaluation dimensions across all papers (Overall) in the manually annotated set (N=278 as 6 out of 284 papers did not report evaluation dimensions), and for the five most frequently occurring NLP task groups. 
    }
    \label{fig:stemmed_dimension}
\end{figure*}

\paragraph{Author-reported tasks and evaluation dimensions}
\label{app:task_dimension_results}
Figure~\ref{fig:stemmed_dimension} presents the distribution of the 15 most frequently evaluated stemmed dimensions across six major NLP task groups. Overall, relevance, coherence, fluency, and correctness dominate the evaluation dimensions across tasks. These dimensions assess whether the generated content is semantically and factually appropriate and related to the task  (e.g., relevance, correctness), and also assess the surface-level linguistic quality (e.g., fluency, coherence). 

Different task groups exhibit different evaluation priorities. In \textit{Dialogue and Interactive Systems}, relevance, coherence, and fluency remain the primary dimensions, accompanied by closer attention to accuracy, while in \textit{Safety and Jailbreak} tasks prioritize relevance, quality, and safety-related dimensions such as correctness and consistency. For \textit{Summarization}, informativeness and faithfulness receive higher emphasis, indicating the importance of content coverage and information consistency. In \textit{Question Answering}, correctness and relevance dominate. \textit{Story generation} places very strong emphasis on coherence. These variations on the evaluated dimensions highlight how evaluation criteria are systematically adapted to the functional goals of different free-form generation tasks.

\section{Bootstrapped Estimates of Proportion}\label{app:bootstrap-table}
We included bootstrapped estimates for reporting frequency for each of our core criteria in Table~\ref{tab:bootstrapped_estimates} along with raw proportions from our manually-annotated sample.
\begin{table*}[th!]
\caption{Bootstrapped estimates (N=500) of proportion of *CL papers that report each of the 20 core criteria, along with the sample proportion (as measured over the manually annotated set).}
\centering
\footnotesize
\begin{tabular}
{p{6.5cm}p{1.4cm}p{2.0cm}p{2.0cm}}
\toprule
Question & Sample Proportion & Bootstrapped Proportion & Bootstrapped Standard Error\\
\midrule
\textbf{Category: Task Documentation} && & \\
\cmidrule{1-4}
Evaluated dimensions reported & 0.9823 & 0.9823 & 0.0003\\
Eval dimensions: justification by prior work & 0.2553 & 0.2542 & 0.0011\\
Eval dimensions: justification provided & 0.5213 & 0.5209 & 0.0013\\
Task guidelines reported & 0.5177 & 0.5169 & 0.0013\\
\midrule
\textbf{Category: Annotation Design} && & \\
\cmidrule{1-4}
Number of annotators reported & 0.7660 & 0.7651 & 0.0011\\
Number of annotated samples reported & 0.8511 & 0.8498 & 0.0009\\
Power analysis used & 0.0000 & N/A & N/A\\
Recruitment platform reported & 0.3511 & 0.3503 & 0.0013\\
Annotator inclusion/exclusion criteria reported & 0.1418 & 0.1420 & 0.0009\\
Method for ensuring quality reported & 0.2163 & 0.2179 & 0.0011\\
Code/annotation interface reported & 0.2801 & 0.2802 & 0.0012\\
Payment information reported & 0.2872 & 0.2872 & 0.0011\\
IRB determination reported & 0.1099 & 0.1099 & 0.0008\\
\midrule
\textbf{Category: Analysis \& Interpretation} && & \\
\cmidrule{1-4}
Annotator demographics reported & 0.7092 & 0.7094 & 0.0012\\
IAA value reported & 0.4610 & 0.4616 & 0.0013\\
IAA sample size reported & 0.1702 & 0.1710 & 0.0010\\
Disagreement resolution method reported & 0.3050 & 0.3057 & 0.0012\\
Data filtering steps reported & 0.0567 & 0.0573 & 0.0006\\
Limitations discussed & 0.1844 & 0.1858 & 0.0010\\
Statistical metric reported & 0.0957 & 0.0973 & 0.0007\\
\bottomrule
\end{tabular}
\label{tab:bootstrapped_estimates}
\end{table*}

\section{Temporal trends}
\label{app:llm-annotated analysis}
\begin{figure*}[th!]
    \centering    
\includegraphics[width=\linewidth]{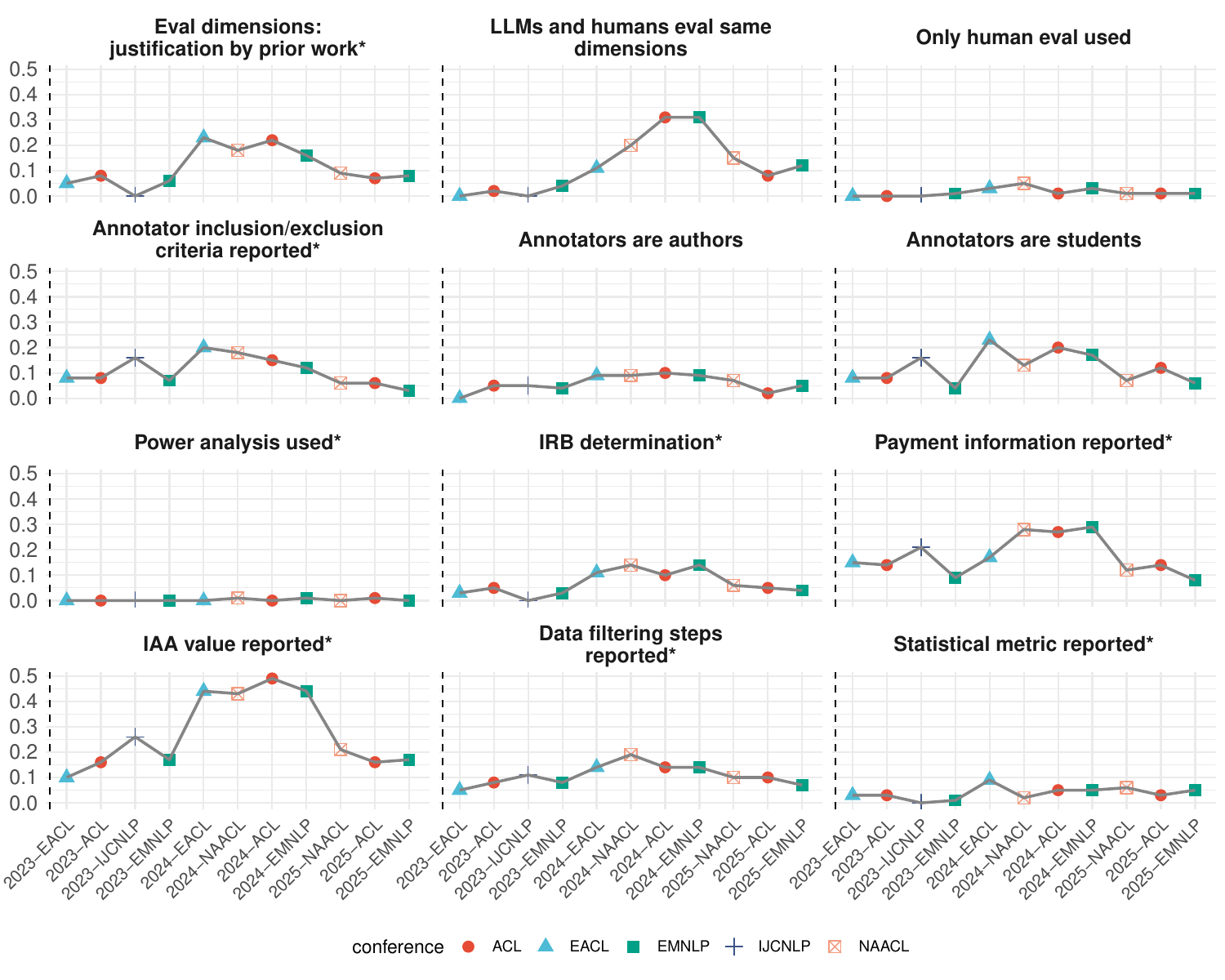}
    \caption{Temporal trends in reporting: across all *CL papers (2023-2025) with human evaluation and long-form generation (N=1,891), frequency of reporting criteria of evaluation protocols remains similar. Notably, we find that the use of LLM-judges is on the rise. Criteria marked with * are among the 20 core reportable criteria.}
    \label{fig:llm_annotation}
\end{figure*}

We provide analysis of temporal trends in reporting based on LLM annotations across *CL papers (2023-2025) with human evaluation and long-form generation (N=1891). As shown in Figure~\ref{fig:llm_annotation}, we observe a similar frequency of reporting criteria of evaluation protocols. However, we also find an increasing adoption of LLM-judges for long-form generation tasks.

\section{Additional analysis}\label{app:additional_analysis}
\begin{figure*}[th!]
    \centering    \includegraphics[width=\linewidth]{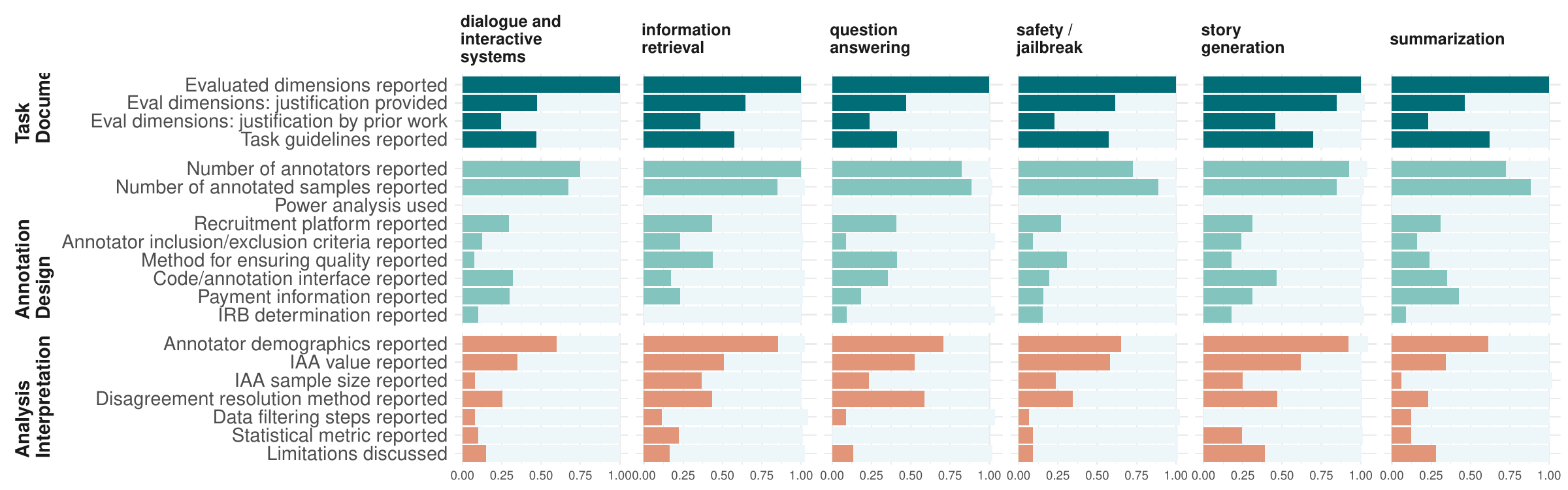}
    \caption{Frequency of Reporting Criteria for Common NLP Tasks}
    \label{fig:reporting_criteria_for_task}
\end{figure*}

\paragraph{Frequency of documenting reportable criteria varies by the most frequent main tasks of the models.}
We provide additional analysis of the breakdown of reportable criteria across different common tasks. As shown in Figure~\ref{fig:reporting_criteria_for_task}, evaluated dimensions, number of annotators, and number of annotated samples are often reported among papers that focus on common tasks. However, reporting for other details related to annotation design and analysis remains infrequent across tasks.

\begin{figure}[th!]
    \centering
    \includegraphics[width=\linewidth]{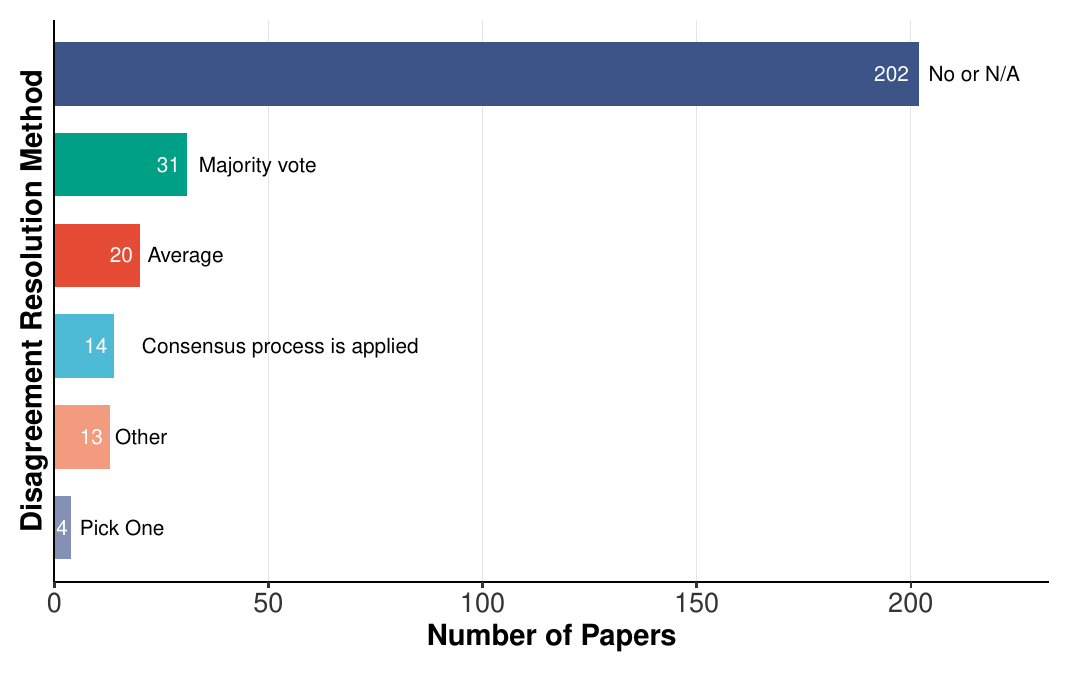}
    \caption{Frequency of disagreement resolution method reported in manually-annotated sample: Most papers tend not to report how they address disagreement among annotators (n=202). Among the ones that report this criteria, majority vote (n=31) is the most common approach for addressing disagreement among annotators, followed by averaging (n=20), consensus process (n=14), other (n=13), or picking one annotation (n=4). }
    \label{fig:disagreement_resolution}
\end{figure}

\paragraph{Frequency of disagreement resolution approaches}
In Figure~\ref{fig:disagreement_resolution}, we include the distribution breakdown of disagreement resolution approaches across the sample of papers we annotated.

\begin{figure}[th!]
    \centering
    \includegraphics[width=1\linewidth]{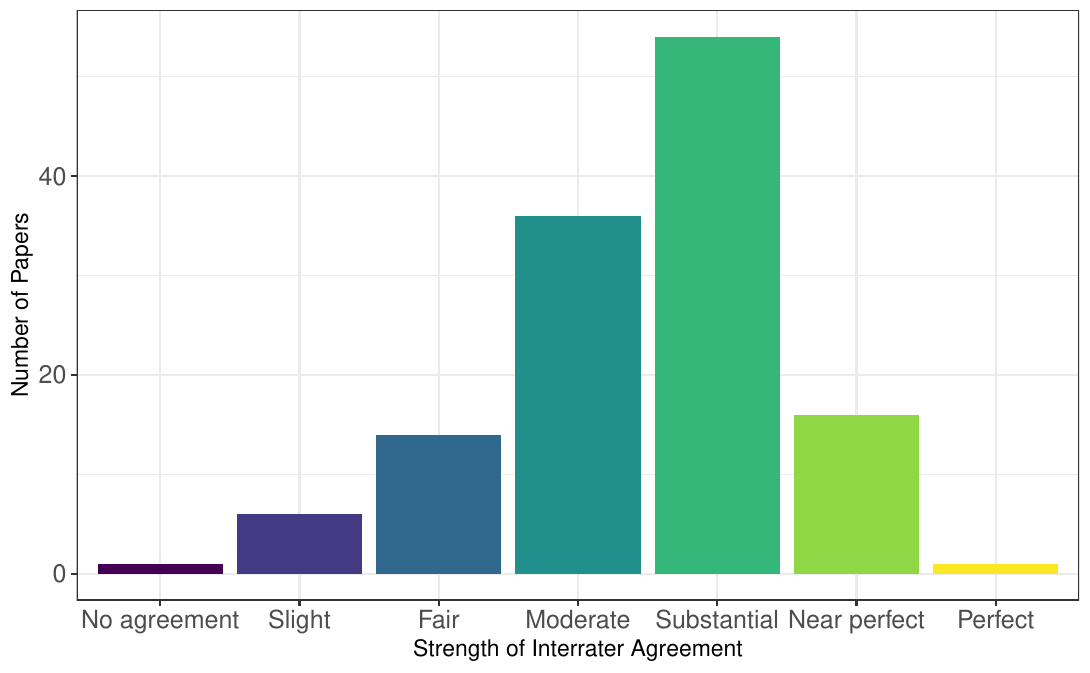}
    \caption{Distribution of IAA strength reported in manually-annotated sample. If an IAA metric value is reported (e.g., Cohen's kappa), we classify the metric value into strength of agreement based on how the metric is usually interpreted.}
    \label{fig: distribution of IAA strength}
\end{figure}

\paragraph{Distribution of IAA}
We provide a detailed breakdown of the strength of interrater agreement reported by our sample. Around 55\% papers did not report the strength of agreement among our annotated sample. Among the papers that reported this information, we find that around 35\% of annotated papers reached moderate agreement and above (see Figure~\ref{fig: distribution of IAA strength}).

\end{document}